\documentclass[oneside,11pt]{article}

\usepackage[backend=biber, style=ieee, isbn=false]{biblatex}
\AtEveryBibitem{
    \clearfield{month}
}

\usepackage{blindtext}

\usepackage{arxiv}
\usepackage{ownstyles}
\usepackage{standalone}

\usepackage[symbol*]{footmisc}
\setfnsymbol{wiley}

\usepackage{lastpage}
\tikzset{
  box/.style={draw, rectangle, minimum size=8mm, font=\small},
  sample/.style={circle, draw, minimum size=8mm, font=\small, inner sep=1pt},
  mini/.style={circle, draw, minimum size=8mm, fill=yellow!30},
  arrow/.style={-Stealth, thick},
}

\newacronym[plural=NNs, firstplural = neural networks (NN)]{NN}{NN}{neural network}
\newacronym[plural=CNNs, firstplural = convolutional neural networks (CNN)]{CNN}{CNN}{convolutional neural network}
\newacronym{GD}{GD}{gradient descent}
\newacronym{losawNN}{losawNN}{local sample weighting neural network}
\newacronym{losaw}{losaw}{local sample weighting}
\newacronym{ML}{ML}{machine learning}
\newacronym{AUC}{AUC}{Area-under-the-curve}
\newacronym{pr-AUC}{pr-AUC}{precision-recall Area-Under-the-Curve}
\newacronym{MSE}{MSE}{mean squared error}
\newacronym{RF}{RF}{random forest}
\newacronym{losawRF}{losawRF}{local sample weighting random forest}
\newacronym{LOCO}{LOCO}{leave-out-covariates}
\newacronym{MDI}{MDI}{mean decrease in impurity}
\newacronym{iid}{i.i.d.}{independently and identically distributed}
\newacronym{SNP}{SNP}{single nucleotide polymorphism}
\newacronym{losawGD}{losawGD}{local sample weighting gradient descent}
\newacronym{ReLU}{ReLU}{Rectified Linear Unit}
\newacronym{LASSO}{LASSO}{least absolute shrinkage and selection operator}

\begin{document}

\title{Decorrelated feature importance from\\ local sample weighting}
\author{
    Benedikt Fröhlich\thanks{University of Regensburg, Faculty of Informatics and Data Science, benedikt.froehlich@ur.de},\quad
    Alison Durst\thanks{University of Regensburg, Faculty of Informatics and Data Science, alison.durst@stud.uni-regensburg.de},\quad
    Merle Behr\thanks{University of Regensburg, Faculty of Informatics and Data Science, merle.behr@ur.de}}

\maketitle

\pagestyle{fancy}

\begin{abstract}
Feature importance statistics provide a prominent and valuable method of insight into the decision process of machine learning models, but their effectiveness has well-known limitations when correlation is present among the features in the training data. In this case, the feature importance often tends to be distributed among all features which are in correlation with the response-generating signal features.
Even worse, if multiple signal features are in strong correlation with a noise feature, while being only modestly correlated with one another, this can result in a noise feature having distinctly larger feature importance score than any signal feature.
Here we propose \textit{\gls{losaw}} which can flexibly be integrated into many \gls{ML} algorithms to improve feature importance scores in the presence of feature correlation in the training data.
Our approach is motivated from inverse probability weighting in causal inference and locally, within the \gls{ML} model, uses a sample weighting scheme to decorrelate a target feature from the remaining features. This reduces model bias locally, whenever the effect of a potential signal feature is evaluated and compared to others.
Moreover, \gls{losaw} comes with a natural tuning parameter, the minimum effective sample size of the weighted population, which corresponds to an interpretation-prediction-tradeoff, analog to a bias-variance-tradeoff as for classical \gls{ML} tuning parameters.
We demonstrate how \gls{losaw} can be integrated within decision tree-based \gls{ML} methods and within mini-batch training of \glspl{NN}. 
For decision trees we integrate \gls{losaw} within the split point selection process to obtain better estimates of the marginal effects of each feature.
For \glspl{NN} we integrate \gls{losaw} within each mini-batch iteration to decorrelate features which are important at the current iteration.
We investigate \gls{losaw} for \gls{RF} and \glspl{CNN} in a simulation study on settings showing diverse correlation patterns.
We found that \gls{losaw} improves feature importance consistently. 
Moreover, it often improves prediction accuracy for out-of-distribution, while maintaining a similar accuracy for in-distribution test data.
\end{abstract}


\section{Introduction}
In recent times, a significant shift of attention within the machine learning (ML) community has moved away from producing mere high-quality prediction models towards focusing on interpretability. This is especially important for biomedical and healthcare science applications of machine learning, where interpretability of \gls{ML} models is often used to deduce from the behavior of the model onto the underlying biological nature, see, e.g., \cite{vellido_importance_2020, lu_importance_2023}.
Arguably one of the most prominent components of interpretable \gls{ML} lies in the construction of feature importance scores, which provide insight into the degree to which a feature influences the model prediction.
There exist many different ways of constructing feature importance scores, including model-specific approaches, such as gradient-based feature importance in neural networks, see, e.g., \cite{morch_visualization_1995}, and mean decrease in impurity for tree-ensemble methods, see, e.g., 
\cite{hastie_elements_2009}, as well as model-agnostic approaches, such as Shapley values, 
permutation-based methods, and \gls{LOCO} methods, see, e.g., \cite{strumbelj_explaining_2014},
\cite{breiman_random_2001}, \cite{lei_distribution_free_2018}.

Feature importance scores have several well-known short-comings, e.g., a general tendency of preferring continuous features for tree-based methods, see, e.g., \cite{strobl_bias_2007, konig_disentangling_2024}, a lack of capturing interaction effects between features, see, e.g., \cite{wright_little_2016}, and general instability problems, see, e.g., \cite{yasodhara_trustworthiness_2021}.
An important scenario where feature importance scores often exhibit undesirable properties, particularly when used to interpret the data-generating process, occurs when features are correlated.
For many prominent feature importance scores, in the presence of correlation the score associated with a true effect feature 
tends to be spread out across all features in correlation to it, see e.g.,
\cite{strobl_conditional_2008, gregorutti_correlation_2017}.
We provide a simple example in the following.

\begin{figure}[h!]
    \centering
    \includegraphics[width=0.8\linewidth]{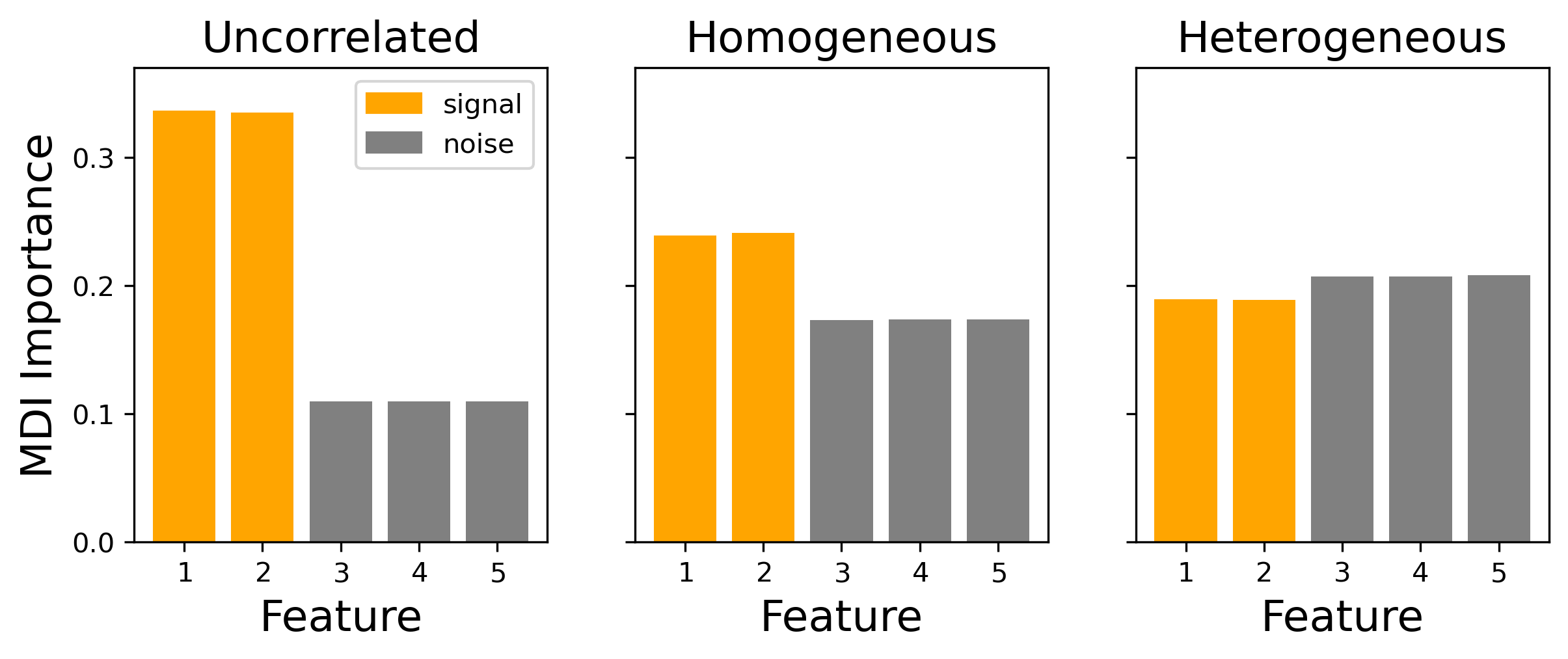}
    \caption{
    Mean decrease in impurity (MDI) feature importance scores of random forest \cite{breiman_classification_1984} averaged over 1,000 Monte Carlo runs for three different feature generation models, as detailed in Example \ref{ex: feature importance for different feature distribution models}. True signal features are shown in yellow, the noise features are shown in gray. Left and middle: When features are independent or when correlation is homogeneous, the true signal features achieve the highest feature importance. Right: For heterogeneous correlation, the noise features achieve the highest feature importance scores.}
    \label{fig: FI barplot for different data distribution}
\end{figure}
\leavevmode
\begin{example}[MDI for \gls{RF} under feature-correlation]
\label{ex: feature importance for different feature distribution models}
We compute \gls{MDI} feature importance scores (\cite{breiman_classification_1984}) for a random forest regressor, where $N = 1,000$ labeled data points, $\{(y^{(n)}, x^{(n)})\}_ {n = 1,\ldots, 1000}$, $x^{(n)}\in \R^5, y^{(n)} \in \R$ with $P = 5$ features, are sampled independently, $(y^{(n)}, x^{(n)}) \sim (Y, X)$, from a linear regression model with $Y = X_1 + X_2 + \epsilon$, for Gaussian additive noise $\epsilon\sim N(0,1)$ which is sampled independently of the features $X$.
The feature distribution is centered multivariate normal $X\sim N(0,\Sigma)$. For the covariance matrix $\Sigma$, we consider three different scenarios, using the identity matrix $I_5$ (uncorrelated features), as well as two matrices $\Sigma_a$ for $a=0.8$ (homogeneous correlation) and $a=0.5$ (heterogeneous correlation), where $\Sigma_a$ is defined by
$$(\Sigma_a)_{i,j}=
\begin{cases}
    1 & i=j \\
    a & \{i,j\}=\{1,2\} \\
    0.8 & i\neq j, \{i,j\}\neq\{1,2\}
\end{cases}.$$

The $m_{try}$ hyperparameter of the \gls{RF} is set to $2=\lfloor P/3\rfloor$.
The resulting feature importance scores, averaged over $1,000$ Monte Carlo runs, are shown in Figure \ref{fig: FI barplot for different data distribution}.
We observe that while for the model with independent features and homogeneous feature correlation \gls{MDI} feature importance is largest for the signal features, in the model with heterogeneous feature correlation the feature importance of noise features is largest.
\end{example}
\leavevmode

\begin{figure}
    \centering
    \includegraphics[width=0.9\linewidth]{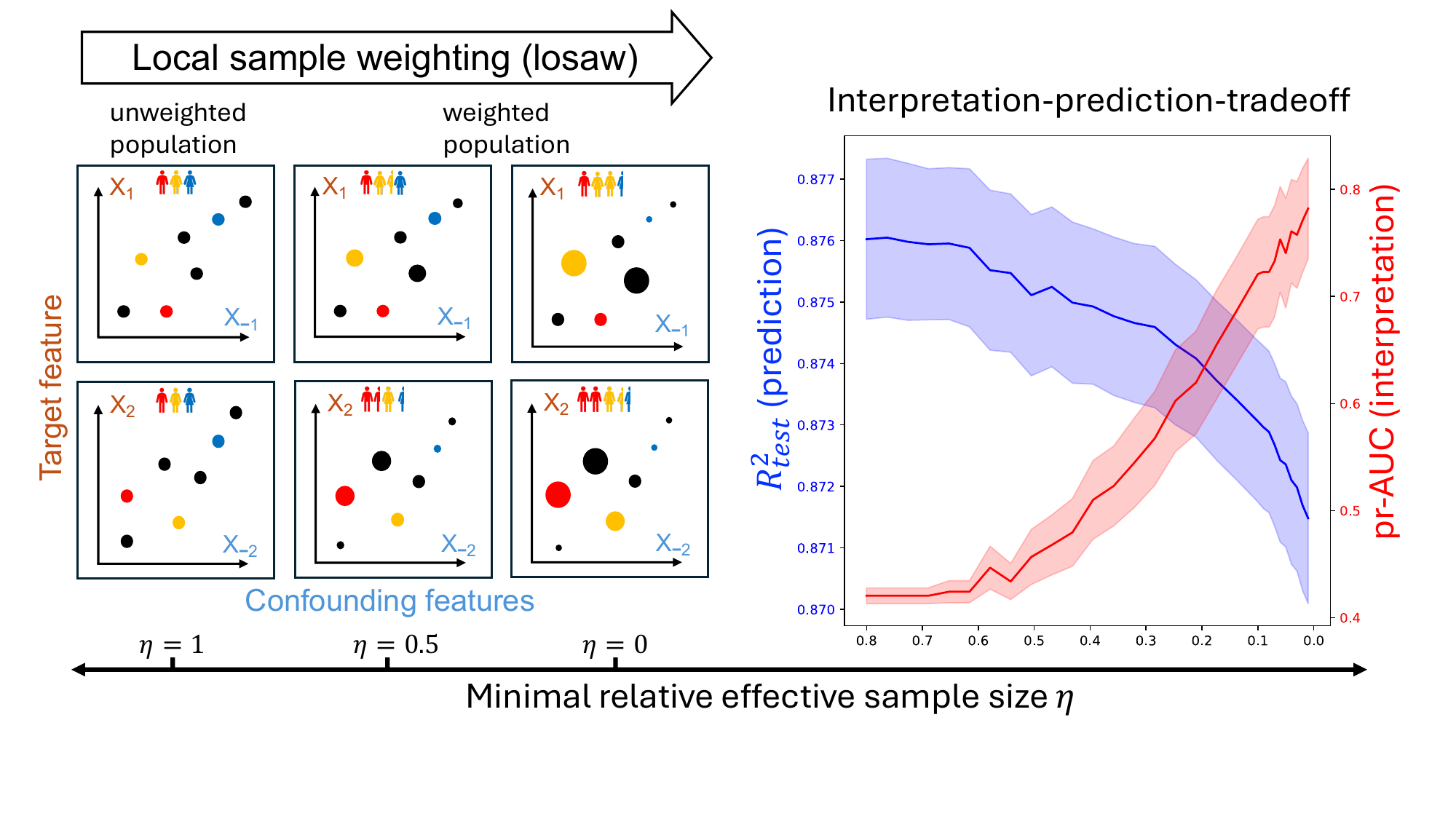}
    \caption{Left: local sample weighting (losaw) with tuning parameter $\eta$: For a given target feature, sample weights are computed such that the target feature becomes independent from the other features (=confounding features) in the weighted population. 
    The largest sample weights are then uniformly redistributed to the remaining samples such that a minimal relative effective sample size of $\eta\in[0,1]$ is ensured. 
    For $\eta=1$, the sample weights are uniform weights. 
    For $\eta=0$, the sample weights are equal to the inverse stabilized propensity scores for the target feature, conditioned on the confounding features.
    Right: Illustration of the interpretation-prediction-tradeoff for $\eta$. Mean test prediction $R^2$ ($R^2_{test}$) and mean \gls{pr-AUC} on the classification of signal features based on feature importance scores for varying relative effective sample size hyperparameter $\eta$ with $95\%$ coverage bands for \gls{losawRF}, see Section \ref{sec: losaw random forest} for details.
    The results show a clear interpretation-prediction-tradeoff with respect to $\eta$.
    }
    \label{fig: illustration for IP-tradeoff}
\end{figure}

\begin{figure}
    \centering
    \includegraphics[width=0.8\linewidth]{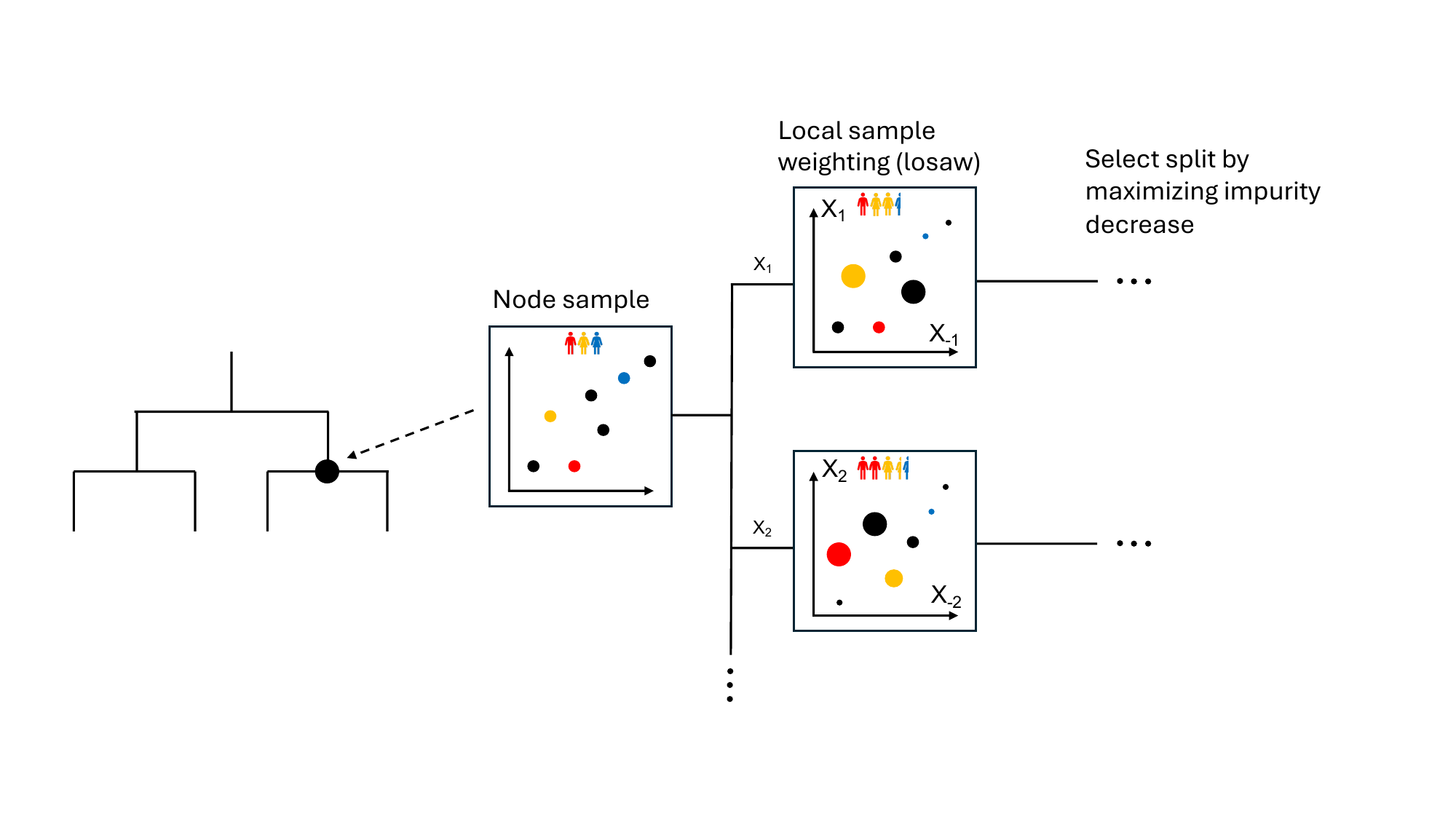}
    \caption{Illustration of \gls{losawRF} on two features. 
    At each node sample, feature-specific sample weights are computed. For each available splitting feature at the node, the corresponding sample weight decorrelates that feature from the remaining covariates. A split is then selected by maximizing impurity decrease across all the re-weighted samples.}
    \label{fig:illustration_losawRF}
\end{figure}

In this paper, we propose a simple, generic approach to correct feature importance scores for feature-correlation, which can be integrated into the training of many \gls{ML} models and combined with many different feature importance scores.
We apply sample weighting to make features uncorrelated in the weighted population and then provide this weighted population to train the \gls{ML} model. 
A similar approach is also used in stable learning, where the aim is to improve prediction accuracy on out-of-distribution samples (\cite{cui_stable_2022, shen_stable_2020-1, shen_stable_2020, yu_stable_2023, yang_stable_2024, kuang_stable_2020}).
However, in stable learning one typically tries to find a global sample weighting scheme which makes all features uncorrelated in the weighted population.
This is statistically challenging and computationally often not feasible, especially in a situation with many correlated features (\cite{kuang_stable_2020, shen_stable_2020-1, yu_stable_2023}).
Here, however, our major goal is not out-of-distribution generalization, but an improvement of feature importance scores. 
Therefore, we do not apply a global weighting scheme, but a local one, where only one target features is considered at a time and the \textit{local sample weighting (\gls{losaw})} is applied in such a way that only the target feature is decorrelated from the remaining features.
If we consider the target feature as some \textit{treatment variable}, for which we want to evaluate its importance, and the remaining features as some \textit{confounding variables}, we can apply standard approaches from inverse propensity score weighting in causal inference (\cite{brown_propensity_2021, robins_marginal_2000}) to obtain sample weights which make the target feature uncorrelated of the remaining features, see the left plot in
Figure \ref{fig: illustration for IP-tradeoff} for an illustration.
As we will show in the following, such an \gls{losaw} approach can flexibly be integrated into the training of many \gls{ML} models, mitigating effects from feature-correlation in the resulting feature importance scores.
In particular, for decision tree-based methods we show how one can successfully use \gls{losaw} at each tree node, to assess the split quality of each feature by decorrelating it from the remaining features, see Figure \ref{fig:illustration_losawRF} for an illustration and Section \ref{sec: losaw random forest} for further details.
For mini-batch training of \glspl{NN} we demonstrate how \gls{losaw} can be applied to individual mini-batches to decorrelate features which currently have a high gradient-based importance in the model from the remaining features, see Figure \ref{fig:losawNN} for an illustration and Section \ref{sec: losaw mini-batch training of neural networks} for further details.

Moreover, we show that one can define a natural tuning parameter for \gls{losaw}, namely, the minimum effective sample size of the weighted population. This tuning parameter gives rise to an \textit{interpretation-prediction tradeoff}, where a large effective sample size results in higher prediction accuracy and a small effective sample size results in better decorrelation and hence higher interpretability, see Figure \ref{fig: illustration for IP-tradeoff} for illustration.

\subsection{Related Work}
The limitations of feature importance statistics on correlated input features are well-known and several approaches to deal with this issue have been proposed. 
In a recent, very related work, adjusted versions of \gls{LOCO} feature importance metrics have been proposed by \cite{verdinelli_decorrelated_2024}, considering a hypothetical product distribution of the feature of interest and remaining features. However, this approach differs to ours, as it is completely model-agnostic, whereas our approach provides a flexible tool that can be integrated within the training of the \gls{ML} algorithm directly. Moreover, 
their approach
is based on one-step estimators using an expression of the respective influence function of the modified LOCO estimator, where as plug-in estimator a full estimator of the joint feature distribution is considered (e.g., derived via kernel density estimators). In contrast, in our approach we consider sample weighting to derive a pseudo-population where a target feature is uncorrelated to the remaining features, which can be integrated into training. In addition, some of the methods proposed in \cite{verdinelli_decorrelated_2024} are restricted to semiparametric models.

In the context of stable learning, features are often separated into stable features that impact the outcome across multiple training distributions, and unstable features, which is closely related to our distinction of signal and noise features. 
Recent publications in this field, e.g., 
\cite{shen_stable_2020, kuang_stable_2020, yang_stable_2024}
and \cite{huang_learning_2023}, have also investigated decorrelation approaches through sample weighting. In the first paper, sample weights are derived from a binary classifier and estimate the density ratio between the data distribution and the product distribution of its marginals, which is closely related to inverse probability weighting, whereas the other three publications compute weights by solving an optimization problem that minimizes sample correlation or saliency-weighted correlation directly. 
Additionally, in \cite{xu_theoretical_2022} the authors derive theoretical properties of sample weighting with respect to out-of-distribution generalization and selection of stable features.
However, their sample weights are computed globally and try to render all features simultaneously independent, as opposed to our approach. In particular, in contrast to our \gls{GD} approach in \gls{NN} training, \cite{huang_learning_2023} do not compute their sample weights locally before each mini-batch and for a particular feature with high importance.
\cite{shen_stable_2020-1} and \cite{yu_stable_2023} employ more targeted weighting schemes that decorrelate only among certain groups of features or among a sparse subset of features, but nonetheless these approaches decorrelate features jointly rather than individually. 
The \gls{RF} alternative introduced by \cite{liao_invariant_2024} requires samples from multiple environments in order to restrict to a set of stable features for its prediction, but cannot learn this stable set from a single environment as opposed to our decorrelation approach.
\cite{zhang_deep_2021} consider sample weights computed for each mini-batch, but their weights are also computed through an optimization problem that decorrelates all features simultaneously. 
It is important to emphasize the fundamental conceptual distinction between stable learning and our approach. While stable learning in the literature typically focuses on improving robustness to covariate shift by decoupling the learning process from unstable features, our method is primarily concerned with improving feature importance scores to identify signal features and enable inference about the underlying data-generating process.

The ability of \gls{ML} models to distinguish between signal and noise features is also investigated in \cite{candes_panning_2018}, but here the authors use knockoff variables to control false discovery of signals, and do not attempt to reduce the correlation of features in the sample. 
A recursive feature elimination algorithm based on permutation importance is studied by \cite{gregorutti_correlation_2017}.
While their approach removes unimportant variables which absorb some of the feature importance, it does not prevent noise features from getting a high importance score in the first place, i.e. it does not discriminate between signal and noise features among highly correlated variables.
\cite{strobl_conditional_2008} introduce a modified version of permutation importance for \gls{RF} to address this problem, but their approach only debiases the feature importance of a \gls{RF} in post-processing but not the learning process itself, and for sufficiently biased splits placed in the learning process, this method may not recover the true signal features from the biased tree prediction model.
A direct integration of debiasing splits into the \gls{RF} training is pursued by
\cite{zhang_novel_2024}. In contrast to our approach, their approach is restricted to binary classification problems and rather than locally decorrelating through sample weighting, their splits are based on estimated causal effect of each feature. 

We also note a certain relatedness between our integration of \gls{losaw} within mini-batch training and the concept of sharpness-aware minimization \cite{clara_training_2025, foret_sharpness-aware_2020, hochreiter_simplifying_1994} in gradient descent \gls{ML} training, where in both cases biased estimates of the gradients are induced. In our case, this is done by the sample weighting and in sharpness aware minimization, this is done by computing an integral of the gradient within a small region to identify flat regions within the gradient landscape.

\subsection{Main contributions}
The main contributions of this paper are as follows:
\begin{itemize}
    \item We propose \textit{\gls{losaw}} as a flexible integration into the training of many \gls{ML} models in order to mitigate negative effects from feature correlation on feature importance scores.
    \item We show that one can define a natural tuning parameter for \gls{losaw}, the minimum effective sample size of the weighted population, which can be related to an \textit{interpretation-prediction tradeoff}, in analogy to a bias-variance tradeoff for classical \gls{ML} tuning parameters.
    \item We propose a new variant of the \gls{RF} algorithm, denoted as \textit{\gls{losawRF}}, together with a respective variant of \gls{MDI} feature importance, where \gls{losaw} is integrated into the local splitting decisions of individual tree nodes.
    \item We propose a new variant of \gls{NN} trained by mini-batch \gls{GD}, denoted as \textit{\gls{losawGD}}, together with a respective variant of gradient-based feature importance, where \gls{losaw} is integrated in the individual training steps.
    \item We demonstrate in a \textit{simulation study} that feature importance from \gls{losawRF} and \gls{losawGD} improve under feature-correlation, compared to the respective state-of-the-art approaches without \gls{losaw}. Moreover, our simulation study shows that our proposed \gls{losawRF} algorithm also often improves in terms of out-of-distribution prediction accuracy, while achieving comparable results in terms of in-distribution prediction accuracy. 
\end{itemize}
\subsection{Notation}
\label{sec: notation}
Throughout this paper, unless specified otherwise, we will denote the training data used to fit our respective \gls{ML} algorithms as $\{(y^{(n)}, x^{(n)})\}_{ n = 1,\ldots, N}$, where $x^{(n)} \in \R^P$ denotes the $P$ different features, $y^{(n)} \in \R$ the label and $N$ denote the sample size. 
For a positive integer $m$, we write 
$[m]$ for the set $\{1,2,\ldots,m\}$, and given an integer 
$p\in[P]$, the symbol $x_p$ denotes the vector of observations of the $p$-th feature in the sample.
Given a subset $S\subset[P]$, we write 
$(\{x^{(n)}\}_{n=1,\ldots,N})_{\vert S}$ for the restriction of the sample onto features in $S$.
By $X=(X_1,\ldots,X_P)$ and $Y$ we denote random variables, for which we assume that the data is sampled from their respective joint distribution. 
With $X_{-p}$ and $x_{-p}$ we denote the respective random variable and sample vector which contains all but the $p$-th feature.
We use the symbols $\P,\E,\var,\cov$ to denote probability, expected value, variance and covariance respectively.
With $N(\mu,\Sigma)$ we denote the multivariate normal distribution with vector of means $\mu$ and covariance matrix $\Sigma$ and with $\phi_{\mu,\Sigma}$ the corresponding probability density function.

\subsection{Outline of the paper}
The paper is structured as follows; in Section \ref{sec: local sample weighting} we introduce our \gls{losaw} approach in its most general form which is not restricted to a specific \gls{ML} model or specific feature importance scores. 
Our effective sample size tuning parameter is introduced in Section \ref{sec: interpretation-prediction-tradeoff} alongside with the corresponding concept of a tradeoff between interpretation and prediction.
In Section \ref{sec: losaw random forest} we integrate our \gls{losaw} approach into a \gls{RF} predictor and construct a suitable impurity-based feature importance score.
We conduct an extensive simulation study that compares our \gls{losawRF} to a standard \gls{RF} regressor on several evaluation metrics measuring predictive performance and interpretability in Section \ref{sec: losaw RF simulation study}.
Afterwards we describe in Section \ref{sec: losaw mini-batch training of neural networks} how \gls{losaw} can be integrated into the training process of a \gls{NN} using saliency maps. 
This approach is also compared with saliency of a standard \gls{NN} in a simulation study in Section \ref{sec: losawNN simulation study}. 
We summarize our findings in Section \ref{sec: discussion} and provide a discussion on the practical applications and impact of our \gls{losaw} algorithms as well as their limitations.
The \hyperref[app: appendix]{Appendix} contains technical details for our algorithms and proofs of all theoretical results stated in the previous sections. 
Additionally, it contains tables summarizing our complete findings in the simulation studies.

\section{Local sample weighting}
\label{sec: local sample weighting}
In the following we introduce \gls{losaw}, which can be integrated flexibly within the training of many \gls{ML} models. Our motivation for \gls{losaw} is that it improves feature importance scores in such a way that they better reflect importance in the underlying data generating process. 
To this end, we assume that our training data is given by $\{(y^{(n)}, x^{(n)})\}_{n = 1,\ldots, N}$, where individual data points were sampled \gls{iid} from some joint distribution $\P_{Y,X}$, that is,
\begin{align*}
    (y^{(n)}, x^{(n)}) \overset{i.i.d.}{\sim} \P_{Y,X}.
\end{align*}
Moreover, in the following, we assume a regression model with additive, independent noise, i.e., for $(Y,X) \sim \P_{Y,X}$ we have that
\begin{align*}
    Y = f(X_1,\ldots,X_P) + \epsilon,
\end{align*}
with $f(x)=\E[Y\mid X = x]$, $\E[\epsilon\mid X]=0$, and $\epsilon$ independent of $X$.
In particular, the domain of $f$ is given by the domain of $X$.
Moreover, with $\P_X$ we denote the distribution of the features $X$, which we often denote simply by $\P$. 
With $\P_{\epsilon}$ we also denote the distribution of the independent additive noise $\epsilon$.
We will denote the pair $(f,\epsilon\sim\P_{\epsilon})$ as the \emph{response generation model}, which in combination with the feature generation model uniquely defines the data generating distribution $\P_{Y,X}$.
For simplicity, we restrict to regression problems for this paper, but we remark that our approaches admit straight-forward extensions to deal also with classification problems.

Many ML algorithms consider different subsets of training observations $\{(y^{(n)}, x^{(n)})\}_{n \in S}$, $S \subset [N]$, at different times during the training process.
For example, in tree-based methods different subsets $S \subset [N]$ are considered at different inner nodes in the tree. For mini-batch \gls{GD} individual mini-batches correspond to different subsets $S \subset [N]$. 
During the training process, the central idea of \gls{losaw} is to estimate feature-specific sample weights $w^p\in\mathbb{R}_{\geq 0}^{|S|}$, for some feature $p \in [P]$, such that the weighted observations approximate the product distribution $\P_p\otimes\P_{-p}$ on $S$, i.e., where feature $X_p$ is independent of the other features on $S$.
\gls{losaw} focuses on different features $p$, via different weighting schemes $w^p(S)$, at different times during the training process. 
For example, in tree-based methods this arises naturally because every node of a tree splits on a single feature. For mini-batch \gls{GD} we can incorporate focusing on individual features during training by sampling a single feature proportional to its current feature importance. 
The key ingredient of \gls{losaw} is to generate estimates of the sample weights $w^p(S)$.
If the full feature distribution $\P$ was known, achieving this is straight forward via weighting with the respective likelihood ratios. 
We can consider this weighting in analogy to inverse propensity score weighting as applied in causal inference. In this case, the \textit{treatment} of the causal inference setting corresponds to the target feature $x_p$ and the \textit{confounders} correspond to the remaining features $x_{-p}$.
With this analogy, the (population) \gls{losaw} weights are defined as follows.
\begin{definition}[\gls{losaw} population weights]\label{def:losawPop}
    Given a feature generation model $X \sim \P$ and a feature $p\in[P]$, the \emph{stabilized propensity score} of an observation $x$ within the domain of $X$ with respect to the $p$-th feature is given by
    \[
    P_p(x_p,x_{-p})\coloneqq\frac{\P(X_p=x_p\mid X_{-p}=x_{-p})}{\P(X_p=x_p)}
    \]
    and \emph{\gls{losaw} population weights} are defined as
    \[w(x) = w^p(x) = P_p(x_p,x_{-p})^{-1}.\]
\end{definition}

Note that for a given feature generation model $X\sim\P$ and feature $p\in[P]$, the weighted pseudo-population with weights $w^p(x)$, as in Definition \ref{def:losawPop}, follows the distribution $\P_{p}\otimes\P_{-p}$.
To see this, let $\Tilde{\P}$ denote the distribution of the weighted pseudo-population. Then we can compute
    \begin{align*}
        \Tilde{\P}(X=x) 
        &=
        \frac{1}{P_p(x_p,x_{-p})}\cdot\P(X=x) 
        =
        \frac{\P(X_p=x_p,X_{-p}=x_{-p})}{\frac{\P(X_p=x_p\mid X_{-p}=x_{-p})}{\P(X_{p}=x_{p})}}
        \\
        &=
        \P(X_p=x_p)\cdot\P(X_{-p}=x_{-p}) =
        \P_p\otimes\P_{p-}(X=x).
    \end{align*}

In practice the feature distribution $\P$ and hence the weights $w^p(x)$ are not known. Therefore, they have to be estimated from the training data, which is completely analog to the estimation of propensity scores in causal inference.
We can estimate the stabilized propensity score $P_p$ at an observation $x$ and feature $p$ by separately computing estimates $\hat{\P}(X_p=x_p\mid X_{-p}=x_{-p})$ and $\hat{\P}(X_p=x_p)$ and considering the resulting estimated \textit{\gls{losaw} weights} 
\begin{align}
    \hat{w}^p(x) = \hat{\P}(X_p=x_p) / \hat{\P}(X_p=x_p\mid X_{-p}=x_{-p}).
\end{align}

When the feature $X_p$ is discrete, a standard approach to estimate the propensity $P_p(x_p,x_{-p})$ is by fitting a \gls{ML} classification model which predicts $x_p$ from $x_{-p}$ and use the predicted class probability of $x_p$ on the input $x_{-p}$.
Analog, if $X_p$ is continuous, one can fit a \gls{ML} regression model $\hat{f}$ which predicts $x_p$ from $x_{-p}$ and estimate the propensity $P_p(x_p,x_{-p})$ by making some distributional assumptions on the residuals, e.g., assuming a centered normal distribution and set $\hat{\P}(X_p=x_p\mid X_{-p}=x_{-p})\coloneqq
        \phi_{0,\hat{\sigma}_p^2}(x_p-\hat{f}(x_{-p}))$,
where $\hat{\sigma}_p^2$ is given by the empiric variance of the residuals of $\hat{f}$.
Alternatively, for continuous features one can also consider non-parametric density estimation approaches, e.g., kernel density estimation, to estimate the propensity ${\P}(X_p=x_p\mid X_{-p}=x_{-p})$.
For estimation of the stabilization score ${\P}(X_p=x_p)$, in the case that feature $X_p$ is discrete, one can consider the empiric class frequency of $x_p$ in the training sample. 
If $X_p$ on the other hand is a continuous feature, one can use any parametric or non-parametric density estimation approach, e.g., fitting a normal distribution or a more flexible approach such as a kernel density estimator, and use its prediction on the value $x_p$ as the estimate for the stabilization score.
We refer to \cite{robins_marginal_2000, brown_propensity_2021} for further discussions on how to estimate stabilized propensity scores.

\subsection{Motivation for losaw}
\label{subsec: motivation for losaw}
In order to motivate our \gls{losaw} approach we will now introduce a suitable notion of signal and noise features. Afterwards we will show how \gls{losaw} can be used to distinguish signal from noise features based on the marginal effect of each feature in the corresponding weighted sample.

\begin{definition}[signal and noise features]
\label{def: signal features}
    For a regression function $f\colon\Omega\subset\mathbb{R}^P\to\mathbb{R}$ with domain $\Omega$, a feature $p\in [P]$ is called a 
    \emph{signal feature} if there exist values 
    $x,x'\in\Omega$ such that $f(x)\neq f(x')$ and
    $x_q=x'_q$ for all $q\neq p$, but $x_p\neq x'_p$.
    Otherwise, we call it a \emph{noise feature}.
\end{definition}

In other words, signal features are precisely those features, for which a change in only the value of this feature can have an effect on the regression function, whereas noise features represent a direction in the domain of the regression function which does not influence the value of the function. Therefore, any regression function $f$ comes with a decomposition 
$[P]=\signal(f) \sqcup\noise(f)$ of the set of features into disjoint subsets of signal and noise features. 
For example, for a linear regression function with $f(x) = \beta_0 + \sum\limits_{p=1}^{P}\beta_p\cdot x_p$ on the feature domain $\Omega = \mathbb{R}^{p}$ we have that $\signal(f)=\{p\in [P]\mid \beta_p\neq 0\}$.

The \gls{losaw} approach, by definition, aims to transform the feature training distribution $\P$ to a product distribution $\P_p\otimes\P_{-p}$, for a specific $p\in[P]$.
In order to justify this transformation in the context of Definition \ref{def: signal features} for signal and noise features, we define the marginal effect and association functions as follows. 

\begin{definition}[marginal association and effect function]
\label{def: marginal ass/eff function}
    For a feature $p\in [P]$, we denote by 
    $\P_{Y,X_p}$ the marginalization of $\P_{Y,X}$ over all features except $X_p$ and $Y$.
    We define the \emph{marginal association function} as 
        \begin{align*}
        f^{ass}(x;p) = \E_{(Y,X_p)\sim\P_{Y,X_p}}[Y\mid X_p = x] = \E_{X\sim\P}[f(X)\mid X_p=x].
    \end{align*}
Moreover, we define the \emph{marginal effect function} as
    \begin{align*}
        f^{eff}(x;p) = \E_{X\sim\P_p\otimes\P_{-p}}[f(X)\mid X_p=x].
    \end{align*} 
 \end{definition}
Note that when the features are already independent, then the marginal association function and the marginal effect function are always the same.
While \gls{ML} models aim to estimate the regression function $f(x) = \E_{\P_{Y,X}}[Y\mid X= x]$, feature importance scores of \gls{ML} models typically capture some form of marginal effect of feature $X_p$, which for many \gls{ML} models and feature importance scores is closely connected to the behavior of the marginal association function $f^{ass}(x;p)$ from Definition \ref{def: marginal ass/eff function}, see, e.g., discussions on marginal vs. conditional approaches \cite{strobl_conditional_2008, watson_testing_2021} in this context.
For example, for decision tree-based methods, at every inner node the marginal effect function at that node for each individual feature $p$ is estimated by a piecewise constant model with a single jump, and the feature for which said approximation results in the largest decrease in impurity determines the splitting feature of that node and hence, feature importance scores such as 
\glsreset{MDI}\gls{MDI}
(\cite{breiman_classification_1984}).
An intrinsic problem of feature correlation for feature importance scores is that features for which the marginal association function is non-constant, do not necessarily correspond to signal features.
We provide a simple example in the following.

\begin{example}[noise features with non-constant marginal association]
\label{ex: noise features with non-constant marginal association}
Consider two discrete features $X_1,X_2 \in \{0,1\}$ with $\P(X_1 = X_2 = 0) = \P(X_1 = X_2 = 1) = 0.4$  and $\P(X_1 = 1, X_2 = 0) =\P(X_1 = 0, X_2 = 1) = 0.1$ and a regression function $f:\{0,1\}^2 \to \{0,1\}$ such that $f(0,0) = f(0,1) = 0$ and $f(1,0) = f(1, 1) = 1$.
Computing the marginal association function of feature $X_2$, we find that
    \begin{align*}
        f^{ass}(0;2)
        &=
        \E[f(X)\mid X_2=0]=
        0.2,
        \quad
        f^{ass}(1;2)
        =
        \E[f(X)\mid X_2=1]=
        0.8,
    \end{align*}
    and thus the marginal association function $f^{ass}(-;2)$ is not constant.
    But as the value of the regression function does not change with the value of $X_2$ in the regression model, feature $X_2$ is a noise feature in this model. 
\end{example}

In contrast to the marginal association function, it turns out that the marginal effect function will always be constant for a noise feature as in Definition \ref{def: signal features} as the following Lemma shows.

\begin{lemma}[marginal signal features are signal features]
\label{lem: (marginal) signal features}
Let the set of \emph{marginal noise feature} be exactly those features for which the function $f^{eff}(x;p)$ is constant. 
Then every noise feature as in Definition \ref{def: signal features} is also a marginal noise feature.
\end{lemma}

\begin{proof}
    We note that since $X_p$ and $X_{-p}$ are independent in the product distribution, we may write
    \begin{align*}
        f^{eff}(x;p) &=
        \E_{X\sim\P_p\otimes\P_{-p}}[f(X)\mid X_p=x] 
        =
        \E_{X_{-p}\sim\P_{-p}}[f(X_p=x,X_{-p})].
    \end{align*}
    If $X_p$ is a noise feature, then for any values $x,x'$ of $X_p$ and $x_{-p}$ of $X_{-p}$ we get the equation
    $f(X_p=x,X_{-p}=x_{-p})=f(X_p=x',X_{-p}=x_{-p})$ and consequently
    \begin{align*}
        f^{eff}(x;p) &=
        \E_{X_{-p}\sim\P_{-p}}[f(X_p=x,X_{-p})] 
        =
        \E_{X_{-p}\sim\P_{-p}}[f(X_p=x',X_{-p})] 
        =
        f^{eff}(x';p).
    \end{align*}
\end{proof}

Lemma \ref{lem: (marginal) signal features} shows that if our data features followed a product distribution of $\P_p$ and $\P_{-p}$ instead of the original feature distribution $\P$, the marginal association of a noise feature with the target vanishes, which shows how \gls{losaw} can
 erase spurious feature importance of noise features in the situation where they are correlated with signal features.

In addition, there also exist signal features which have a constant marginal association function, but a non-constant marginal effect function, as shown in the following example. This indicates that \gls{losaw} may also identify some signal features which are not detected otherwise.

\begin{example}[signal features with constant marginal association and non-constant marginal effect]
\label{ex: signal features with constant marginal association}
Consider a regression function $f$ with feature distribution $\P$ as in Example \ref{ex: noise features with non-constant marginal association}, i.e., $\P(X_1 = X_2 = 0) = \P(X_1 = X_2 = 1) = 0.4$  
and $\P(X_1 = 1, X_2 = 0) =\P(X_1 = 0, X_2 = 1) = 0.1$. As a regression function consider $f:\{0,1\}^2 \to \{0,1\}$ such that $f(0,0) = 0, f(0,1) = -3, f(1,0) = 5$, and $f(1, 1) = 2$. 
Feature $X_2$ is a signal features in this case. However, the marginal association function $f^{ass}(-;2)$ is constant, while this is not the case for the marginal effect function $f^{eff}(-;2)$. 
\end{example}

\begin{remark}[signal features with constant marginal effect]
\label{rem: signal features with constant marginal association and effect function}
We note that signal features can have both, a constant marginal association and a constant effect association function, even when all features are independent. This can happen due to specific non-linear effects in the regression function $f(x)$. A classical example, which is also often given for a situation where decision tree-based methods fail, is given by two independent features that can both take values in $\{0,1\}$, each with equal probability $0.5$, and where the regression function is given by XOR.
While such a situation can result in poor performance of feature importance scores, the core reason for this problem is in the regression function itself and not in the feature correlation structure. 
\end{remark}

\section{Interpretation-Prediction-Tradeoff}
\label{sec: interpretation-prediction-tradeoff}
Before we provide detailed examples on how \gls{losaw} can be incorporated into the training of \gls{ML} models and respective feature importance scores, we highlight the interpretation-prediction-tradeoff which follows from a \gls{losaw} approach more generally.
Any non-uniform sample weighting will lead to a reduced effective sample size.
Especially when high correlation is present in the data, and consequently some of the rare observations have rather small conditional probabilities, the corresponding inverse probability weights of those observations can become very large, causing them to dominate the weighted sample.
This can be problematic, because an estimator will essentially be trained on only these observations, resulting in a significantly reduced effective sample size.
It is therefore crucial to quantify the sample size associated with all weighting schemes in our algorithm and if necessary, modify these weights in a way which guarantees that the sample size remains reasonable even after re-weighting.
A popular and easy-to-compute estimate of effective sample size is given by the \emph{Kish formula} below.
\begin{definition}[effective sample size (Kish formula \cite{kish_survey1966})]
    For a sample weight $w\in\R_{\geq 0}^N$, its \emph{effective sample size} is defined as
    \[
    s(w)\coloneqq\frac{\Vert w\Vert_1^2}{\Vert w\Vert_2^2}=
    \frac{(\sum\limits_{n}w_n)^2}{\sum\limits_{n}w_n^2},
    \]
    and takes values in $(0,N]$. We also define the \emph{relative effective sample size} 
    $s_{rel}(w)\coloneqq\frac{s(w)}{N}\in (0,1]$ quantifying the effective sample size of the weighted sample relative to the unweighted sample.
\end{definition}
Now we can measure the effective sample size associated with a sample weight, but we also need some way of modifying weights in a way which guarantees our desired effective sample size.
Given a normalized sample weight $w\in\R_{\geq 0}^N$ and a real threshold $\theta\geq\frac{1}{N}$, we iteratively set all weights equal to or exceeding the threshold to $\theta$ and distribute the excess weight uniformly among the remaining weights, until no weight exceeds the threshold anymore.
We denote the weights obtained by this algorithm by $w_\theta$. 
    Given an effective sample weight threshold $\eta\in (0,1]$, we define 
\begin{align}
     \tilde{\theta}(\eta)\coloneqq\max\{\theta\in[1/N,1]\mid s_{rel}(w_{\theta})\geq\eta\}.
\end{align}
    
That is, the largest threshold $\theta$ for which the modified weight $w_{\theta}$ has relative effective sample size $\geq\eta$.
For a proof of the convergence of this redistribution algorithm as well as for details on the binary search algorithm which we use to approximate the threshold $\tilde{\theta}$ we refer to the appendix.
The parameter $\eta \in (0,1]$ is a tuning parameter of the \gls{losaw} approach.
A large $\eta$ corresponds to only a small weighting of the samples, hence, the effective sample size is not reduced much, however the features are also not decorrelated well. Hence, this corresponds to a high prediction accuracy (due to large effective sample size) but low interpretability (due to strong feature correlation). 
On the other hand, a small $\eta$ potentially corresponds to a small effective sample size and hence, a low prediction accuracy, but a good decorrelation of features and a high interpretability.

The computation of such a modified sample weight from estimated stabilized propensity scores for a given effective sample size tuning parameter is performed using the Algorithm \ref{alg: pseudocode for weight modification algorithm}. 
For more details, we refer to the Algorithms 
\ref{alg: enforce maximal weight threshold} and \ref{alg: binary search for weight} in the Appendix.

\begin{Algorithm}
\label{alg: pseudocode for weight modification algorithm}

    \begin{algorithm}
        \begin{algorithmic}\\
            \State\textbf{Inputs:} Estimated stabilized propensity scores $\hat{P}_{p}\in\R_{>0}^N$, minimal relative effective sample size tuning parameter $\eta\in[0,1]$, tolerance $
            \alpha>0$
            \State\textbf{Output:} 
            Inverse probability sample weights $w$, modified to guarantee a relative effective sample size $\eta$ (up to tolerance $\alpha$).
            \State
            \State $w\gets 1/\hat{P}_p$
            \Comment{Inverse stabilized propensity scores}
            \State $w\gets w/\Vert w\Vert_1$
            \Comment{Normalize sample weight}
            \If{$s_{rel}(w)\geq\eta$}
            \Comment{sufficient relative effective sample size}
            \State\Return $w$
            \EndIf
            \State\Return Modified weight $w_{\theta}$ satisfying $s_{rel}(w_\theta)\in[\eta-\alpha,\eta+\alpha]$ as computed by the Algorithm 
            \ref{alg: enforce maximal weight threshold} for $\theta\in\left[\frac{1}{N\eta},1\right]$ according to Algorithm \ref{alg: binary search for weight}.
        \end{algorithmic}
    \end{algorithm}
\end{Algorithm}

As an example, we consider Random Forest \cite{breiman_random_2001} with \gls{losaw} incorporated together with a respective version of \gls{MDI} feature importance, see Section \ref{sec: losaw random forest} for details.
To illustrate the tradeoff between interpretation and prediction, we compute \gls{pr-AUC} for the binary classification task predicting the signal features based on the feature importance scores, as a  measure for \textit{interpretability}, and test prediction $R^2$, as a measure of \textit{predictability}, see Section \ref{sec: losaw RF simulation study} for more details. 
The tuning parameter $\eta$ is varied from $0.01$ to $0.80$ with $200$ Monte Carlo runs for each $\eta$. Each sample consists of $N = 1,000$ i.i.d. observations with $P = 10$ features from the linear regression model
    \[
    X\sim N(0,\Sigma),\quad Y = X_1 + X_2 + \epsilon,\quad 
    \epsilon\sim N(0,1),
    \]
    with covariance matrix containing both homogeneous and heterogeneous correlation, recall Example \ref{ex: feature importance for different feature distribution models}, given by
    \[
    \Sigma = 
    \left[
    \begin{array}{@{}c|c|c@{}c|c|c|c}
         \begin{matrix}
            1 & 0.4 & 0.8 \\
            0.4 & 1 & 0.8 \\
            0.8 & 0.8 & 1
         \end{matrix}
         &
         0.2
         & 
         0
         \\ \hline
         0.2
         & 
         \begin{matrix}
             1 & 0.9 & 0.9 \\
             0.9 & 1 & 0.9 \\
             0.9 & 0.9 & 1
         \end{matrix}
         &
         0 \\ \hline
         0
         &
         0
         &
         I_{4}
    \end{array}
    \right].
\]
The test $R^2$ is then computed using an independent test sample of size $1,000$ from the same regression model.
The true signal features for the regression function are given by $X_1$ and $X_2$.
The respective mean $R^2$ and mean \gls{pr-AUC} can be seen in the right plot of Figure 
\ref{fig: illustration for IP-tradeoff}, which clearly depicts the interpretation-prediction-tradeoff: by lowering the tuning parameter $\eta$, we can greatly increase the \gls{pr-AUC} and trade off a small amount of predictive performance for a strong increase in terms of \gls{pr-AUC}, causing the feature importance scores to more accurately reflect the true signal features.

\section{losaw random forest}
\label{sec: losaw random forest}
In the following, we provide the details on how \gls{losaw} can be integrated into the \gls{RF} algorithm, which we denote as \gls{losawRF}.
\gls{losawRF} is identical for most parts with the standard implementation of a \gls{RF} regressor, see \cite{breiman_random_2001, hastie_elements_2009} for details on \gls{RF}, in general. Its main difference lies in the way in which it selects a pair of splitting feature and splitting point, which is done by maximizing a \emph{weighted relative impurity decrease}. 

\begin{definition}[weighted relative impurity decrease]
\label{def: rel imp dec}
    Given a training dataset \\
    $\{(y^{(n)},x^{(n)})\}_{n=1,\ldots,N}$ together with a normalized sample weight $w\in\R^N_{\geq 0}$, we denote by 
    \[
    \overline{y}_w\coloneqq\sum\limits_{n=1}^{N}w_n y^{(n)},\quad
    \MSE_w\coloneqq\sum\limits_{n=1}^{N}w_n{y^{(n)}}^2-\overline{y}_w^2
    \]
    the \emph{weighted prediction} and \emph{weighted mean squared error} associated with the weight $w$. 
    If moreover $x$ is a split point for the $p$-th feature, we denote by 
    \[
    w^{-}(x,p)\coloneqq\sum\limits_{n=1}^N\one(x^{(n)}_p\leq x)\cdot w_n,\quad
    w^{+}(x,p)\coloneqq\sum\limits_{n=1}^N\one(x^{(n)}_p>x)\cdot w_n
    \]
    the \emph{total weight} in the left and right childe node.
    Assuming they are both positive, we can furthermore define weighted predictions
    \begin{align*}
        \overline{y}_w(x,p)^{-}
        &\coloneqq
        \frac{1}{w^{-}(x,p)}\sum\limits_{n=1}^{N}\one(x_p^{(n)}\leq x)\cdot w_n\cdot y^{(n)}
        \\
        \overline{y}_w(x,p)^{+}
        &\coloneqq
        \frac{1}{w^{+}(x,p)}\sum\limits_{n=1}^{N}\one(x_p^{(n)}> x)\cdot w_n\cdot y^{(n)}
    \end{align*}
    for both child nodes using the respective renormalized weights.
    Likewise, we can define weighted mean squared errors 
    \begin{align*}
        \MSE_w(x,p)^{-}
        &\coloneqq
        \left(\frac{1}{w^{-}(x,p)}\sum\limits_{n=1}^{N}\one(x_p^{(n)}\leq x)\cdot w_n\cdot{y^{(n)}}^2\right)-{\overline{y}_w(x,p)^{-}}^2,
        \\
        \MSE_w(x,p)^{+}
        &\coloneqq
        \left(\frac{1}{w^{+}(x,p)}\sum\limits_{n=1}^{N}\one(x_p^{(n)}> x)\cdot w_n\cdot{y^{(n)}}^2\right)-{\overline{y}_w(x,p)^{+}}^2
    \end{align*}
    for both of the child nodes. Given these, we can now define the \emph{weighted impurity decrease} 
    \[
    \Delta_w(x,p)\coloneqq \MSE_w - w^{-}(x,p)\cdot\MSE_w^{-} -
    w^{+}(x,p)\cdot\MSE_w^{+}
    \]
    and
    \emph{relative weighted impurity decrease}
    \begin{align}\label{eq:relDelta}
         \Delta_w^{rel}(x,p)\coloneqq\frac{\Delta_w(x,p)}{\MSE_w}
    \end{align}
    for a split at the $p$-th feature and with splitting value $x$.
    Moreover, whenever $w^{-}(x,p)=0$ or $w^{+}(x,p)=0$, we define 
    $\Delta_w^{rel}(x,p)\coloneqq\Delta_w(x,p)\coloneqq 0$.
\end{definition}

\begin{definition}[split selection criterion]
\label{def: split selection criterion}
    Given training data $\{(y^{(n)},x^{(n)})\}_{n=1,\ldots,N_k}$ at a node $k$, a set of available splitting features $M_{try}\subset [P]$ and a tuple of normalized sample weights 
    $w=(w_1,\ldots,w_P) \in \R_{\geq 0}^{N \times P}$, the pair $(\tilde{p},\tilde{x})$ of splitting feature $\tilde{p}\in M_{try}$ and splitting value $\tilde{x}\in\{x_p^{(n)}\}_{n=1,\ldots,N_k}$ selected by \gls{losawRF} is given by
    \[
    (\tilde{p},\tilde{x})\coloneqq\argmax\limits_{p\in M_{try},\ x\in(x_p^{(1)},\ldots,x_p^{(N)})}\Delta_{w_p}^{rel}(x,p).
    \]
\end{definition}

Pseudo-code for this split-selection  of \gls{losawRF} is given in Algorithm \ref{alg: pseudocode split selection algorithm}.
\begin{Algorithm}[split selection for \gls{losawRF}]
\label{alg: pseudocode split selection algorithm}

\begin{algorithm}
    \begin{algorithmic}\\
        \State\textbf{Input:} 
        A node dataset $\{(y^{(n)},x^{(n)})\}_{n=1,\ldots,N_k}$, a set of features 
        $M_{try}=\{p_1,\ldots,p_{m_{try}}\}\subset[P]$, tuning parameter $\eta\in (0,1]$ for the minimal relative effective sample size.
        \State\textbf{Output:}
        The optimal splitting feature $\tilde{p}\in M_{try}$ and splitting point $\tilde{x}\in (x_{\tilde{p}}^{(1)},\ldots,x_{\tilde{p}}^{(N_k)})$
        \For{$i=1,\ldots,m_{try}$}
        \State 
        Estimate the stabilized propensity scores
        $\{\hat{P}_{p_i}(x_{p_i}^{(n)},x_{-p_i}^{(n)})\}_{n=1,\ldots,N_k}$ 
        \State from $\{x^{(n)}\}_{n=1,\ldots,N_k}$ 
        as in Section \ref{sec: local sample weighting}
        \State 
        Compute sample weights 
        $\hat{w}_i$ from $\hat{P}_{p_i}$ with a relative effective sample size $\geq\eta$ 
        \State as in Algorithm \ref{alg: pseudocode for weight modification algorithm}
        \State
        Compute 
        $\Delta_{\hat{w}_i}^{rel}\coloneqq\Delta_{\hat{w}_i}^{rel}(x_{p_i}^{(n)},p_i)_{n=1,\ldots,N_k}$
        \State
        $n_i\gets\argmax\limits_{n=1,\ldots,N}\Delta_{\hat{w}_i}^{rel}$
        \State $\tilde{x}_i\gets x_{p_i}^{(n_i)}$
        \Comment{Optimal splitting point for feature $p_i$}
        \State $\Delta_i\gets\Delta_{\hat{w}_i,n_i}^{rel}$
        \Comment{Maximize relative weighted impurity decrease for feature $p_i$}
        \EndFor
        \State 
        $\tilde{i}\gets\argmax\limits_{i=p_1,\ldots,p_{m_{try}}}\Delta_i$
        \Comment{Maximize relative weighted impurity decrease over all features}
        \State\textbf{return }
        $(\tilde{p}=p_{\tilde{i}},\ \tilde{x}=\tilde{x}_{\tilde{i}})$
    \end{algorithmic}
\end{algorithm}
\end{Algorithm}

Note that in the standard implementation of RF there is no weighting and hence maximizing the relative impurity decrease as in (\ref{eq:relDelta}) is equivalent to maximizing the impurity decrease directly. However, for losawRF the weighted mean squared error $\MSE_{w_p}$ is not constant across features, but depends upon the sample weight $w_p$ associated with a feature.
If we do not correct for this locality, we expect the split selection algorithm to be biased towards features where the weighted impurity in the parent node is higher, as then a comparably small relative reduction in weighted impurity can be associated with a large absolute weighted impurity decrease.
In an extreme case, we may think of an example with two sample weights $w_1,w_2$ such that a split of feature $1$ at a value $x$ results in an absolute impurity decrease 
    $\Delta_{w_1}(x,1)>\MSE_{w_2}$
    that exceeds already the total weighted impurity in the parent node for feature $2$. 
    Then, even if some split at feature $2$ may result in total purity, a split selection rule based on absolute weighted impurity decrease would prompt us to split on feature $1$, rather than feature $2$.
    The rescaling by the total weighted impurity therefore establishes a fair comparison across all features.

\begin{remark}
    As the weighted impurity at the parent node $\MSE_w$ does not depend on the split point, we may factor out the rescaling factor to rephrase the optimization problem as follows.
    \begin{align*}
        (\tilde{p},\tilde{x})
        &\coloneqq
        \argmax\limits_{p\in M_{try},\ x\in (x_p^{(1)},\ldots,x_p^{(N_k)})}\Delta_{w_p}^{rel}(x,p)
        =
        \argmax\limits_{p\in M_{try},\ x\in(x_p^{(1)},\ldots,x_p^{(N_k)})}\frac{1}{\MSE_{w_p}}\Delta_{w_p}(x,p)
        \\
        &=
        \argmax\limits_{p\in M_{try}}\frac{1}{\MSE_{w_p}}\Delta_{w_p}
        \left(\argmax\limits_{x\in(x_p^{(1)},\ldots,x_p^{(N_k)})}\Delta_{w_p}(x,p),p\right)
    \end{align*}
    Therefore, we can first compute for each feature in $p\in M_{try}$ the split point maximizing the weighted (unscaled) impurity decrease, rescale all of the associated impurity decreases by $1/\MSE_{w_p}$ and then select the $p$ maximizing this value.
    There are well-known efficient algorithms computing the optimal split point for a given pair $(p,w_p)$ both for the case of discrete and continuous features. 
    For a detailed description, we refer to the Appendix \ref{app: split point selection}.
\end{remark}

\subsection{MDI feature importance for losawRF}
\label{subsec: feature importance for losawRF}
Mean decrease in impurity (MDI) is one of the standard implementations for random forest feature importance, see \cite{hastie_elements_2009}.
It is computed for each tree estimator by summing up for each feature the individual impurity decreases at the nodes, weighted with the size of the node sample, and then normalizing these scores to sum up to $1$. These tree-specific feature importances are then averaged to obtain a feature importance for the tree-ensemble.
For \gls{losawRF} we use essentially the same metric constructed from the node-specific relative weighted impurity decreases, but this time also weighted with the target variance of the respective node training sample.

\begin{definition}[feature importance]\label{def: feature importance}
    Let $T$ be a single tree in a \gls{losawRF} consisting of $m_T$ inner nodes which are indexed by $k=1,\ldots,m_T$. 
    For each inner node $k$, we denote by $p_k$ its splitting feature, and by $\Delta_k^{rel}$ the weighted relative impurity corresponding to the node. 
    For the training sample associated with the node, we denote the variance of the responses as $\MSE_k$ and its sample size with $N_k$.
    We then define an \emph{unstandardized impurity score}
    $\FI^{T,pre}\in\R^P$ of the tree $T$ as
     \[
    FI^{T,pre}_p\coloneqq\sum\limits_{k=1}^{m_T}\one(p_k=p)\cdot\Delta_k^{rel}\cdot MSE_k\cdot N_k,
    \]
    and a \emph{standardized impurity score}
    $FI^T\coloneqq FI^{T,pre}/\Vert FI^{T,pre}\Vert_1$ for the tree $T$.
    For the entire \gls{losawRF} consisting of trees 
    $T_1,\ldots,T_{n_{tree}}$, we define its vector of \emph{feature importances} by averaging the individual tree-specific standardized feature importance scores
    \[
    FI\coloneqq\frac{1}{n{_{tree}}}\sum\limits_{t=1}^{n_{tree}}FI^{T_t}.
    \]
\end{definition}

The additional factor of $\MSE_k$ in Definition \ref{def: feature importance} ensures that this feature importance metric ranges on the same scale as the standard \gls{MDI} feature importance for RF. 
The feature importance score for \gls{RF} weights each impurity decrease $\Delta_k$ (which lies in the interval $[0,\MSE_k]$) at a node $k$ with the node sample size $N_k$. Therefore the contribution of each node lies in the interval $[0,N_k\cdot\MSE_k]$. 
Since the weighted relative impurity decrease $\Delta_k^{rel}$ is measured proportionally to the absolute weighted impurity, it lies by definition in the interval $[0,1]$, causing the contribution of each node to lie in $[0,N_k]$ after weighting it with the node sample size.
So by multiplying all relative weighted impurity decreases at nodes with the respective node sample $\MSE$, we enforce that the individual contributions to the feature importance metrics are comparable to the standard implementation of random forest MDI.

\subsection{Computational complexity of losawRF}
\label{subsec: generalizations}
So far we described \gls{losawRF} in its most general form, modifying only the split selection process for a given set of feature-specific sample weights, which renders the computational complexity highly dependent on the way in which these weights are obtained.
In order to provide explicit complexity classes we therefore restrict our attention now to two explicit methods with which these weights are computed, differentiating between the case of discrete and continuous features. 
    We assume that for the propensity estimation, a subset of $Q$ 
    \emph{adjustment features} has been selected from the set of all features, and that the machine learning estimator computing the propensity scores only takes into account these $Q$ variables as features, see Remark \ref{rem: adjustment features}.
    Additionally, we assume that for the case of discrete and continuous features, the following estimators are used to compute the propensity scores and stabilization scores.
    \begin{enumerate}
        \item \textbf{For discrete features.} The propensity scores are estimated by fitting a multinomial logistic regression model predicting the propensity feature from the set of adjustment features. 
        The stabilization scores are estimated using frequency count.
        \item \textbf{For continuous features.} The propensity scores are estimated from the residual error of a linear regression model predicting the propensity feature from the set of adjustment features. 
        The stabilization scores are estimated from the density function of a normal distribution with mean and variance estimated from the \emph{full} training sample (i.e., not just the node sample).
    \end{enumerate}
\begin{proposition}[time complexity classes for \gls{losawRF}]
\label{prop: time complexity classes for losawRF}
    Given a training dataset with $P$ features and sample size $N$, we fit a \gls{losawRF} with $m_{tree}$ tree estimators which considers $m_{try}=O(P)$ features for each node and that uses $Q$ adjustment features for the propensity estimation. 
    For discrete features, we assume each feature can take up to $K$ different values.
    Assuming that each child node contains at least a proportion of $\gamma$ observations compared to its parent, for some constant $\gamma\in (0,0.5]$, 
    the \gls{losawRF} has a training complexity of 
    \begin{enumerate}
        \item $O(m_{tree}PQN\log (N)K^2)$ for discrete features,
        \item $O(m_{tree}PN\log(N)(Q+\log N))$ for continuous features.
    \end{enumerate}
\end{proposition}

We refer to the Appendix for details on an efficient implementation of the \gls{losawRF} algorithm realizing these complexity classes 
as well as a detailed proof of Proposition \ref{prop: time complexity classes for losawRF}.
Ignoring the additional $\log N$ factors, this differs from the computational complexity of a standard random forest implementation by a factor of $Q$, with $Q \leq P$. 

\begin{remark}[adjustment features]
\label{rem: adjustment features}
    A reasonable selection of the $Q$ adjustment features is important to guarantee computational efficiency when the number of features $P$ is large.
    In practice, 
    we propose selecting the adjustment covariates by training an initial \gls{ML} model on the data and using its top $Q$ most important features, assuming that they cover the signal features. 
    Additionally, we assume that the propensity estimation for each feature relies mostly on those features which are in high correlation with it, 
    so we restrict the adjustment features further by filtering our features whose absolute correlation with the respective splitting feature does not pass a certain threshold.
\end{remark}

\subsection{Simulation study for losawRF}
\label{sec: losaw RF simulation study}
In order to compare the performance of a \gls{RF} regressor to our \gls{losawRF} algorithm, we simulate a broad variety of data types, correlation structures and response generation models for varying sample size, dimension and noise level, assessing the performance of both algorithms and the quality of their feature importance metrics with identical hyperparameter settings on the respective evaluation metrics.

\paragraph{Feature generation model.}
We want to generate both, samples with \emph{continuous} and \emph{discrete} features whilst capturing a large variety of correlation structures and different dimension and sample sizes. To keep the total number of simulation scenarios manageable, we make the following assumptions.
\begin{itemize}
    \item all features follow the same marginal distribution: a standard normal distribution in the case of continuous features, and a centered binomial distribution 
    with $2$ trials and success rate $0.5$ for the case of discrete features. 
    \item features $X_7,X_8,\ldots,X_p$ are sampled independently from the others from their respective marginal distribution, while the first six features are sampled from a joint distribution, which is either
    a multivariate normal distribution $N(0,\Sigma)$ with a correlation matrix $\Sigma$ (in particular, each feature follows a standard normal distribution), or in the discrete case 
    a joint distribution 
    $\P_{joint}$ approximating the correlation matrix 
    $\Sigma$.
    \footnote{The joint distribution $\P_{joint}$ is computed using numerical optimization, see Appendix \ref{app: discrete distribution with numerical optimization}. 
    }
\end{itemize}
The choice of the correlation matrix $\Sigma$ should reflect various interesting types of correlation behavior within the features, in particular: (i) blocks of correlated features with little correlation in between blocks, (ii) a block of features with high, homogeneous correlation, and (iii) a block of features with heterogeneous correlation.
Using the following correlation matrix for our simulation, we can satisfy all of these three requirements.
\[
\Sigma = 
\left[
\begin{array}{@{}c|c@{}}
     \begin{matrix}
        1 & 0.4 & 0.8 \\
        0.4 & 1 & 0.8 \\
        0.8 & 0.8 & 1
     \end{matrix}
     &
     0.2
     \\ \hline
     0.2
     & 
     \begin{matrix}
         1 & 0.9 & 0.9 \\
         0.9 & 1 & 0.9 \\
         0.9 & 0.9 & 1
     \end{matrix}
\end{array}
\right]
\]
Blocks of heterogeneously and homogeneously correlated features are given by the feature blocks $X_1,X_2,X_3$ and $X_4,X_5,X_6$, respectively.

\paragraph{Response generation model.}
We assume that the response $Y$ is generated from the covariates $X$ following a regression function $f$ and independent Gaussian noise $\epsilon$,
\[
Y = f(X) + \epsilon,\quad \epsilon\sim N(0,\sigma^2).
\]
To ensure uniform signal-to-noise ratios across different choices for the regression function $f$, we set 
$\sigma^2\coloneqq\phi\cdot\hat{\var}(f(X))$ for a parameter 
$\phi$ controlling the signal-to-noise ratio which is varied from a setting of low noise $(\phi=0.1)$ and a setting of high noise ($\phi = 1.0)$ across simulations.
The variance $\var(f(X))$ of the signal is estimated using a sample of size $10,000$ drawn from the same feature generation model as the training data.
The regression functions used in the simulations should reflect a diverse range of possible regression behavior, including linear effects and constant effects which are activated at a given threshold, as well as interacting features and different constellations between effect features and feature correlation. 
We use the following seven regression models as summarized by the Table \ref{tab: simulation regression models}, alongside with their corresponding set of signal features.
\begin{table}[ht]
    \centering
    \caption{Regression models used in the simulation study, together with their regression function and set of signal features following Definition \ref{def: signal features}.}
    \begin{tabular}{|c|l|l|}\hline
         \multicolumn{1}{|c}{Regression model} &  
         \multicolumn{1}{|c|}{Regression function} & 
         \multicolumn{1}{c|}{Signal features} \\
         \hline
         $1$ & $f_1(X) = X_4$ & $\{X_4\}$ \\
         $2$ & $f_2(X) = X_1 + X_4$, & $\{X_1,X_4\}$ \\
         $3$ & $f_3(X) = X_1 + X_2$, & $\{X_1,X_2\}$ \\
         $4$ & $f_4(X) = X_1 + X_2 + X_4$ & $\{X_1,X_2,X_4\}$ \\
         $5$ & $f_5(X) = \one(X_1\geq 0)\cdot\one(X_2\geq 0)$ & $\{X_1,X_2\}$ \\
         $6$ & $f_6(X) = \one(X_1\geq 0)\cdot\one(X_4\geq 0)$ & $\{X_1,X_4\}$\\
         $7$ & $f_7(X) = \one(X_1\geq 0)\cdot\one(X_2\geq 0)+\one(X_4\geq 0)$ & $\{X_1,X_2,X_4\}$ \\ \hline
    \end{tabular}
    
    \label{tab: simulation regression models}
\end{table}

The first four regression functions consider linear models. Among them, the first two models either consist of the relatively easy cases of either a single signal, or two signal features in two distinct correlation blocks. 
Models $3$ and $4$ concern the more difficult cases where the linear effect features are moderately correlated with one another, but in high correlation with the same noise feature, and model $4$ also contains a distinct linear effect from a different block.
The remaining three models $5,6,7$ resemble models $3,2$ and $4$ but instead consider an interaction effect of two features, activated by a threshold on both features, plus an additional threshold effect in a different correlation block in the case of model $7$, which correspond to so-called locally-spiky-sparse (LSS) models, see \cite{basu_iterative_2018,behr_lss_2022}.

\paragraph{Simulation parameters.}
In each simulation, we specify the data and response generation model by their 
\emph{data type} (discrete or continuous),
\emph{sample size} $N$ ($=500$ or $5000$),
\emph{dimension} $P$ ($=10,50$ or $100$),
\emph{signal-to-noise ratio} $\phi$ ($=0.1$ or $1.0$) and
\emph{regression function} $f$ ($=f_1$,... or $f_7$).
For each of these parameter settings, we draw $250$ Monte Carlo samples from the respective feature/response generation models and use them to train a \gls{losawRF} and a \gls{RF} regressor.

\paragraph{Algorithm specifications.}

For the machine learning models used for the estimation of propensity scores $\hat{\P}(X_p=x_p\mid X_{-p}=x_{-p})$, we use  multinomial logistic regression in the case of discrete features, and linear regression for the case of continuous features.
In both cases, up to $Q=10$ adjustment features are selected for the propensity estimation model
based on \gls{MDI} importance of a \gls{RF} model and a correlation threshold of $0.1$, as described in 
Remark \ref{rem: adjustment features}.
For the estimation of stabilizing scores in the case of a continuous feature $X_p$, we use a normal distribution.
After computing the corresponding inverse probability weights, these weights are modified to guarantee a minimal relative effective sample size of at least $\eta = 0.25$ using Algorithm \ref{alg: pseudocode for weight modification algorithm}.
Both, the \gls{RF} regressor implemented by the scikit-learn python package 
(see \cite{scikit-learn}) and \gls{losawRF}, use $100$ individual decision trees for their prediction which are fit to a tree-specific bootstrap dataset drawn from the original training dataset and with the same sample size.
Moreover, all trees are constructed to a maximal depth of $10$ and with minimal leaf sizes of $5$ data points. 
The $m_{try}$ hyperparameter controlling the number of available splitting features at each tree nodes is set to $\lfloor P/3\rfloor$ for both algorithms. 

\paragraph{Evaluation metrics}
In each Monte Carlo run of the simulation, a test dataset of sample size $N=1,000$ is generated using the same regression model as the training data.
Additionally, we generate a second dataset using a modified regression model, which admits the noise term from the response generation model and where the feature generation model is replaced by the product of the marginal distributions.
In this second dataset the features are therefore independent but follow the same marginal distribution as for the training dataset.
For both algorithms we compute the test prediction $R^2$ on these two datasets.
To evaluate performance with respect to the feature importance scores, we consider a \gls{pr-AUC} statistic.
More precisely, this is defined as follows.
Let $FI\in\R_{\geq 0}^P$ denote a vector of feature importance scores associated to a machine learning model fit to the training data.
Depending on a feature importance cut-off $\theta$, we can define a binary classifier on the set $[P]$ of features, associating the label ``signal" to any feature $p$ which satisfies $FI_p\geq\theta$, and ``noise" to the remaining features.
Denoting by $\signal_\theta$ and $\noise_\theta\subset[P]$ the features classified as signal or noise, we may compute 
    \[
    \text{precision}_\theta\coloneqq\frac{\signal(f)\cap\signal_\theta}{\signal_\theta},\quad
    \text{recall}_\theta\coloneqq\frac{\signal(f)\cap\signal_\theta}{\signal(f)}
    \]
as the proportion of properly identified signal features with respect to all features classified as signals or with respect to all true signal features.
In particular, varying the threshold $\theta$ we can compute a precision-recall curve based on feature importance scores, and its \gls{AUC}, measuring how well the feature importance separates signal from noise features.

\paragraph{Results of simulation study}
Our findings indicate that depending on the regression function, \gls{losawRF} either greatly outperforms the classical \gls{RF} regressor in terms of \gls{pr-AUC} and $R^2$ on independent features or achieves at least highly comparable results on these metrics. 
These improvements are stable across varying dimension $P$, sample size $N$ and noise level $\phi$ for both discrete and continuous features. 
Simultaneously, the $R^2$ on an independent test set for \gls{losawRF} is highly comparable to the one achieved by 
\gls{RF}, with the largest difference being equal to $0.018$.
An overview of the simulation results for the settings with $P=100$ features and low noise level $\phi=0.1$ can be found in Table \ref{tab: results continuous} for the case of continuous features, and in Table \ref{tab: results discrete} for the case of discrete features. 
We refer to the Appendix \ref{app: all results} for a full account of all simulation results.

The greatest improvements can be found for the regression functions $f_3,f_4,f_5$ and $f_7$ which contain at least two signal features within a heterogeneous correlation block. 
Specifically for the case of discrete features and sample size $N=5,000$, we can observe that \gls{losawRF} achieves a near perfect \gls{pr-AUC} of $0.999$ and $0.961$ for the regression functions $f_3$ and $f_4$, 
while the corresponding values of \gls{RF} range at $0.417$ and $0.514$, indicating an improvement of $0.582$ and $0.445$ on this score.
These observations show that whereas the feature importance scores of \gls{RF} are not sufficient to correctly identify the signal features, for our local sample-weighting random forest algorithm they serve as reliable separators between signal and noise features.
For the remaining regression functions $f_1,f_2$ and $f_6$ there is either only one signal feature, or the signal features are contained in two distinct correlation blocks. 
In these three settings, both algorithms achieve nearly perfect \gls{pr-AUC} scores and in terms of $R^2_{ind}$ there is no observable tendency for either algorithm to outperform the other.
With the exception of one scenario, the scores achieved by \gls{losawRF}
were always better or comparable (difference $<0.05$) to the results of \gls{RF}. These exceptions occurred for the case of $P=100$ discrete features, high noise ($\phi=1.0$) and low sample size ($N=500$), where the $R^2_{ind}$ score for the regression function $f_2$ achieved by \gls{losawRF} was $0.078$ smaller than for RF.
From this simulation study we draw the conclusion that on scenarios with multiple correlated signal features, \gls{losawRF} was able to consistently outperform \gls{RF}, while maintaining a similar performance on the remaining scenarios. 
\begin{table}
    \centering
    \caption{Simulation results for \gls{RF} and \gls{losawRF} for $P=100$ continuous features and noise parameter $\phi=0.1$ for all seven regression functions.
    The evaluation metrics are averaged over all $250$ Monte Carlo runs and are given by test prediction $R^2$ for an independent test dataset ($R^2_{test}$) and on a dataset with independent features and zero noise ($R^2_{ind}$), as well as \gls{pr-AUC} for the classification of signal features using feature importance scores.
    The algorithm which performs best in terms of the respective metric is marked in boldface, the underlined numbers indicate that one algorithm outperformed the other by more than $0.05$ on the given metric.
    }
    \begin{tabular}{|c|l||c|c|c|c|}\hline
        regression & evaluation & \multicolumn{2}{|c|}{$N = 500$} & \multicolumn{2}{|c|}{$N=5000$} \\ \cline{3-6}
        function & metric & RF & losawRF  & RF  & losawRF  \\ \hline\hline
        \multirow{3}{*}{\hfil$1$}  
        & $R^2_{test}$  & \B{0.871} & 0.869  & 0.901 & \B{0.901} \\
        & $R^2_{ind}$   & 0.686 & \B{0.723}  & 0.806 & \B{0.822} \\
        & pr-AUC        & 1.000  & 1.000 & 1.000 & 1.000  \\   \hline
        \multirow{3}{*}{$2$}  
        & $R^2_{test}$  & \B{0.830} & 0.823  & \B{0.889} & 0.887 \\
        & $R^2_{ind}$   & 0.681 & \B{0.709}  & 0.823 & \B{0.849} \\
        & pr-AUC        & \B{1.000} & 0.999  & 1.000 & 1.000 \\   \hline
        \multirow{3}{*}{$3$}  
        & $R^2_{test}$  & \B{0.849} & 0.845  & 0.885 & \B{0.888} \\
        & $R^2_{ind}$   & 0.315 & \U{\B{0.516}}  & 0.445 & \U{\B{0.668}} \\
        & pr-AUC        & 0.417 & \U{\B{0.547}} & 0.417 & \U{\B{0.656}} \\   \hline
        \multirow{3}{*}{$4$}  
        & $R^2_{test}$  & \B{0.833} & 0.823  & \B{0.879} & 0.877 \\
        & $R^2_{ind}$   & 0.369 & \U{\B{0.520}}  & 0.464 & \U{\B{0.685}} \\
        & pr-AUC        & 0.513 & \U{\B{0.659}}  & 0.514 & \U{\B{0.714}} \\ \hline
        \multirow{3}{*}{$5$}  
        & $R^2_{test}$  & 0.830 & \B{0.848}  & 0.902 & \B{0.903} \\
        & $R^2_{ind}$   & 0.752 & \U{\B{0.815}}  & \B{0.972} & 0.969 \\
        & pr-AUC        & 0.629 & \U{\B{0.959}} & 0.881 & \U{\B{1.000}} \\   \hline
        \multirow{3}{*}{$6$}  
        & $R^2_{test}$  & 0.844 & \B{0.850}  & 0.902 & \B{0.903} \\
        & $R^2_{ind}$   & 0.826	 & \B{0.846}  & \B{0.979} & 0.970 \\
        & pr-AUC        & 1.000 & 1.000 & 1.000 & 1.000 \\   \hline
        \multirow{3}{*}{$7$}  
        & $R^2_{test}$  & 0.829 & \B{0.843}  & 0.898 & \B{0.901} \\
        & $R^2_{ind}$   & 0.741 & \B{0.805}  & 0.949 & \B{0.964} \\
        & pr-AUC        & 0.734 & \U{\B{0.958}}  & 0.919 & \U{\B{1.000}} \\   \hline
    \end{tabular}
    \label{tab: results continuous}
\end{table}
\begin{table}
    \centering
    \caption{Simulation results for \gls{RF} and \gls{losawRF} as in Table \ref{tab: results continuous}, but for discrete features
    }
    \begin{tabular}{|c|l||c|c|c|c|}\hline
        regression & evaluation & \multicolumn{2}{|c|}{$N = 500$} & \multicolumn{2}{|c|}{$N=5000$} \\ \cline{3-6}
        function & metric & RF & losawRF & RF & losawRF \\ \hline\hline
        \multirow{3}{*}{\hfil$1$}  
        & $R^2_{test}$ &  0.880 & \B{0.881}  & 0.904 & \B{0.904} \\
        & $R^2_{ind}$ & 0.705 & \B{0.713}  & 0.945 & \B{0.953} \\
        & pr-AUC & 1.000 & 1.000 & 1.000 & 1.000 \\   \hline
        \multirow{3}{*}{$2$}  
        & $R^2_{test}$ & \B{0.874} & 0.867  & 0.897 & \B{0.898} \\
        & $R^2_{ind}$ & \B{0.718} & 0.703  & 0.907 & \B{0.923} \\
        & pr-AUC & 1.000 & 1.000 & 1.000  & 1.000 \\   \hline
        \multirow{3}{*}{$3$}  
        & $R^2_{test}$ & 0.885 & \B{0.886} & 0.898 & \B{0.902} \\
        & $R^2_{ind}$ & 0.265 & \U{\B{0.358}}  & 0.600 & \U{\B{0.787}} \\
        & pr-AUC & 0.417 & 0.417 & 0.417 & \U{\B{0.999}} \\   \hline
        \multirow{3}{*}{$4$}  
        & $R^2_{test}$ & \B{0.864} & 0.863  & 0.897 & \B{0.901} \\
        & $R^2_{ind}$ & 0.306 & \U{\B{0.378}}  & 0.592 & \U{\B{0.799}}\\
        & pr-AUC & 0.485 & \B{0.510} & 0.514 & \U{\B{0.961}} \\   \hline
        \multirow{3}{*}{$5$}  
        & $R^2_{test}$ & 0.868 & \B{0.871}  & 0.906 & \B{0.908}\\
        & $R^2_{ind}$ & 0.656 & \B{0.693}  & 0.899 & \U{\B{0.959}}  \\
        & pr-AUC & 0.571 & \U{\B{0.797}}  & 0.702 & \U{\B{1.000}} \\   \hline
        \multirow{3}{*}{$6$}  
        & $R^2_{test}$ & 0.873 & \B{0.875} & 0.906 & \B{0.907} \\
        & $R^2_{ind}$ & 0.780 & \B{0.810} & 0.961 & \B{0.972}\\
        & pr-AUC & 1.000 & 1.000  & 1.000 & 1.000 \\   \hline
        \multirow{3}{*}{$7$}  
        & $R^2_{test}$ & \B{0.879} & 0.879  & 0.905 & \B{0.906} \\
        & $R^2_{ind}$ & 0.695 & \B{0.738} & 0.871 & \U{\B{0.932}} \\
        & pr-AUC & 0.696 & \U{\B{0.797}}  & 0.797 & \U{\B{1.000}} \\   \hline
    \end{tabular}
    
    \label{tab: results discrete}    
\end{table}

\section{losaw mini-batch gradient descent}
\label{sec: losaw mini-batch training of neural networks}
In the following, we describe how \gls{losaw} can be integrated into mini-batch gradient descent (GD) training, as commonly used, e.g., in the training of a \gls{NN}. For a comprehensive overview of mini-batch GD see, e.g., \cite{sgd}. We refer to this integrated approach as losawGD.
The primary distinction between \gls{losawGD} and standard implementations of \gls{GD} lies in the mini-batch sampling strategy. Whereas standard mini-batches are drawn uniformly without replacement across training steps, \gls{losawGD} samples mini-batches according to dynamically assigned weights over the sample and with replacement. 
More specifically, for a current iteration $d$ of a \gls{GD} step, a specific feature $X_p(d), p \in [P]$, is selected and the sample weights are obtained via the \gls{losaw} approach as outlined in Section \ref{sec: local sample weighting}. The weights are then used to sample the observations for the mini-batch \gls{GD}.
To determine the feature $X_p(d)$ which specifies the \gls{losaw} weights, feature importance scores $FI(d) \in \R_+^P$ are calculated at each iteration $d$ and $X_p(d)$ is sampled proportional to $FI(d)$. 
The losawGD approach is illustrated in Figure \ref{fig:losawNN}. 
In our implementation, see Algorithm \ref{alg:losawNN} and Section \ref{sec: losawNN simulation study}, we consider training of a \gls{NN} with \gls{losawGD}, where feature importance is quantified via the saliency method by \cite{Simonyan14a}. However, the \gls{losawGD} framework is compatible with arbitrary feature importance metrics and, more generally, with any learner that can be trained by \gls{GD}. 
\begin{remark}
    While standard implementations of \gls{GD} sample the observation of a mini-batch without replacement, in \gls{losawGD} the observations of the mini-batches in each training step are sampled with replacement from the weighted population. Hence the same observation may be included multiple times within a mini-batch and across successive training steps, which renders the epoch hyperparameter obsolete and replaces it by the total number of mini-batch training steps $D$. 
\end{remark}
\begin{figure}
    \centering 
    \scalebox{0.8}{
    \usetikzlibrary{positioning, shapes, arrows.meta, calc}
\definecolor{myc}{rgb}{0.4,0.6,1}

\tikzset{
  box/.style={draw, rectangle, minimum size=8mm, font=\small},
  sample/.style={circle, draw, minimum size=8mm, font=\small, inner sep=1pt},
  mini/.style={circle, draw, minimum size=8mm, fill=yellow!30},
  arrow/.style={-Stealth, thick}}%
\begin{tikzpicture}[node distance=0mm and 0mm]

  \foreach \i/\y/\col/\val in {
    1/0.0/c0/0.63,
    2/1.0/c1/0.4,
    3/2.0/c2/0.98,
    4/3.0/c3/0.11,
    5/4.0/c4/0.21
  }{
    \pgfmathsetmacro{\pct}{100*\val}
    \pgfmathtruncatemacro{\ipct}{\pct}
    \node[box, fill=myc!\ipct!white] (T\i) at (0,-\y) {\val};
    \node[left=2pt of T\i] {$X_\i$};
  }
    \node[above=5pt of T1] {$FI(d)$};
    
    \draw[arrow] (T3.east) -- ++(0,0mm) -- ++(2,0) coordinate (sel);

     \node[box, fill=myc!98!white, right=2cm of T3](Feat) {0.98};

\node[] (samples) at (7,-2) {};
\node[draw, rounded corners, inner sep=2mm] (samplesbox) at (7,-2) {
  \begin{tikzpicture}[node distance=2mm]
    \foreach \idx/\val in {
      1/0.01, 2/0.11, 3/0.51, 4/0.09,
      5/0.78, 6/0.34, 7/0.26, 8/0.91,
      9/0.07,10/0.21,11/0.17,12/0.54
    }{
      \pgfmathsetmacro{\pct}{100*\val}
      \pgfmathtruncatemacro{\ipct}{\pct}
      \pgfmathtruncatemacro{\row}{int((\idx-1)/4)}
      \pgfmathtruncatemacro{\col}{mod((\idx-1),4)}
      \node[sample, fill=myc!\ipct!white]
            (s\idx) at (\col*1,-\row*1) {\val};
    }
  \end{tikzpicture}
};
\node[above=2.55cm of samples]{losaw};
\node[above=2.15cm of Feat] {$X_p(d)$};
  \draw[arrow] (Feat.east)  -- (samplesbox.west);
   
   \node[](minibatch) at (12,-2) {};
   \node[above=2.4cm of minibatch]{$\mathcal{B}(d)$};
 \node[draw, rounded corners, inner sep=2mm] (mini) at (12,-2) {
    \begin{tikzpicture}[node distance=2mm]
    \node[mini, fill=myc!26!white] (m1) {};
    \node[mini, right=of m1,fill=myc!78!white] (m2) {};
    \node[mini, below=of m1, fill=myc!91!white] (m3) {};
    \node[mini, right=of m3, fill=myc!54!white] (m4) {};  
    
  \end{tikzpicture}};

  \draw[arrow] (samplesbox.east)   -- (mini.west);
    
    \node[coordinate] (corner0) at ($(mini.east)+(0.5,0cm)$){};
    \node[coordinate] (corner4) at ($(corner0)+(0,3.5cm)$){};
    \node[coordinate] (corner1) at ($(corner4)+(-15.4,0)$){};
    \node[coordinate] (corner2) at ($(corner1)+(0,-3.5cm)$){};
    \node[coordinate] (corner3) at ($(T3.west)+(-0.7,0cm)$){};

    \draw[arrow, thick, black] (mini.east)--
					(corner0) -- (corner4) -- (corner1) node[midway, sloped, above, text=black] {\small Update model parameters in {GD} iteration $d$} -- (corner2) -- (corner3);

    \node[align=center] at (-0.5,-5.5) {\small Compute $FI(d)$\\  for each feature.};
    \node[align=center] at (3,-5.5) {\small Sample one feature $X_p(d)$ \\based on $FI(d)$.};
    \node[align=center] at (7,-5.5) {\small Reweight samples \\ for target feature $X_p(d)$.};
    \node[align=center] at (12,-5.5) {\small Draw mini-batch $\mathcal{B}(d)$\\ based on sample weights.};
    \path[use as bounding box] ([shift={(0.0,0.2)}]current bounding box.south west) --
                              ([shift={(1.0,-0.2)}]current bounding box.north east);

\end{tikzpicture}}
    \caption{Gradient descent training with \gls{losaw}, depicted schematically. 
    In each gradient descent step $d$, \gls{losawGD} computes saliency-based feature importance scores $FI(d)$ and samples a feature $X_p(d)$ with probability proportional to these scores.
    For this feature, inverse stabilized propensity score weights are estimated. 
    Then a mini-batch sample is drawn proportionally to these sample weights, which is used for \gls{GD} update of the model parameters.
    }
\label{fig:losawNN}
\end{figure}
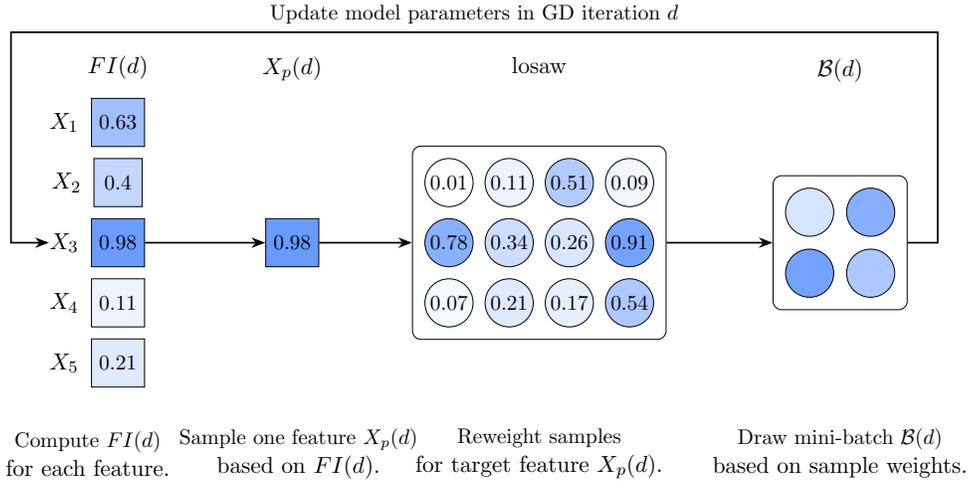
\begin{Algorithm} [losawGD with saliency]
\label{alg:losawNN}
\begin{algorithm}
    
    \begin{algorithmic}\\
        \State\textbf{Input:} 
        A training dataset $\{(y^{(n)},x^{(n)})\}_{n=1,\ldots,N}$, number of training steps $D$, size of mini-batch $B$, initialized model parameters $\hat{\beta}$, minimal relative effective sample size $\eta$.
        \State\textbf{Output:}
        Learned weights $\hat{\beta}$ of \gls{NN}. 
            \For{$d=1,\ldots,D$} \Comment {Training step loop}
                \For {$p=1,\ldots,P$} \Comment{Compute saliency}
                    \State $s_p \gets \frac{1}{N}\sum_{i=1}^N \lvert \frac{\partial}{\partial x_p} f_{\hat{\beta}}(x^{(i)})\rvert_{\text{min}=0}^{\text{max}=1}$
                    
                \EndFor
                \State $s\gets(s_1,\ldots, s_P)$
                \State $s \gets s/{\sum_{p=1}^P s_p}$                 
                \State Draw a feature $p\in[P]$ with probabilities $s_1, \ldots, s_P$   
                
                \State Estimate stabilized propensity scores $\{\hat{P}_{p}(x^{(n)}_p,x^{(n)}_{-p})\}_{n=1,\ldots,N}$ \Comment{see Section \ref{sec: local sample weighting}}
                \State Compute sample weights $\{\hat{w}_n\}_{n=1,\ldots,N}$ from $\hat{P}_{p}$ with $\eta$     \Comment{see Appendix \ref{app: effective sample size}}    
                \State Draw a mini-batch $\mathcal{B}$ of $B$ observations from $\{(y^{(n)},x^{(n)})\}_{n=1,\ldots,N}$ with \newline  \hspace*{1.5cm} probabilities $\hat{w}_1, \ldots, \hat{w}_N$ with replacement
                \State Perform mini-batch gradient descent on $\mathcal{B}$ and update  $\hat{\beta}$
            \EndFor
    \end{algorithmic}
\end{algorithm}
    
\end{Algorithm}

\subsection{Simulation study for losawGD}
\label{sec: losawNN simulation study}
As for \gls{losawRF}, we conduct a simulation study to compare the performance of a \gls{NN} trained using TensorFlow's standard training loop (see \cite{abadi_tensorflow_2016}) with that of a \gls{NN} trained via our own \gls{losawGD} training loop. We vary correlation structures and response generation models and evaluate the predictive performance and the quality of the final feature importance scores (via saliency maps) of both resulting \glspl{NN}.

\paragraph{Feature generation model.} Since \glspl{NN} are typically applied in settings characterized by a large number of covariates and a high volume of observations, the simulation study is designed accordingly, employing datasets with $P=1,000$ features and $N=50,000$ samples.
Here, we only investigate discrete features with the following assumptions similar to the feature generation model as in Section \ref{sec: losaw RF simulation study}.
\begin{itemize}
    \item All features follow the same marginal distribution, a binomial distribution with two trials and success rate 0.5. 
    \item All features are samples from a joint distribution $\P_{joint}$ approximating a correlation matrix of the form of $\Sigma$, as detailed below (see Appendix \ref{app: discrete distribution with numerical optimization}).
\end{itemize}
The choice of correlation matrix $\Sigma$ is more limited in the high dimensional and discrete setting, due to the computational complexity of approximating the correlation matrix $\Sigma$. Here, we build it from a $(5\times 5)$-block of fixed correlation structure $\sigma$, given by
\begin{align*}
    \sigma = 
\left[
\begin{array}{@{}c}
     \begin{matrix}
        1 & \alpha & \beta & \alpha & \beta\\
        \alpha & 1 & \beta & \beta & \beta\\
        \beta & \beta &1& \beta & \beta\\
        \alpha & \beta & \beta& 1 & \beta\\
         \beta & \beta &\beta  & \beta&1\\
     \end{matrix}
\end{array}
\right].
\end{align*}
for some randomly selected values of $\alpha, \beta$, with $\alpha>0.7$ and $\beta<0.3$, see Appendix \ref{app: discrete distribution with numerical optimization} for details.
Again, this correlation matrix design ensures that we have both homogeneous and heterogeneous correlation within the features.
To obtain the final correlation matrix $\Sigma$, one block $\sigma$ is randomly generated and concatenated to obtain the feature correlation matrix $\Sigma$ of the form
\begin{align*}
   \Sigma = 
\left[
\begin{array}{@{}c}
\begin{matrix}
     \sigma &  &\bigzero      \\
     & \ddots\\ 
     \bigzero &  & \sigma   
\end{matrix}   
\end{array}
\right].
\end{align*}
More details on the feature generation is provided in Appendix \ref{app: discrete distribution with numerical optimization}.

\paragraph{Response generation model.} The same assumptions on the regression function as in Section \ref{sec: losaw random forest} apply for this simulation study. In this simulation study, we only consider the setting of high noise ($\phi=1.0$), with estimated variance $\var(f(X))$ by using a sample of 10,000 observations drawn from the same data generation model as the training data. We use three regression models as summarized by Table \ref{tab: simulation regression models NN}, where in each case the signal features are given by $\{X_s\mid s\in \mathcal{S}\}$ with $\mathcal{S}:=\{1,3,201,204,601,602,603\}$. This choice of signal features ensures that a combination of independent, heterogeneous, and homogeneous correlation structures are present in the regression problems and allow for different combination of interactions in regression function $f_{10}$.

\begin{table}[ht]
    \centering
    \caption{Regression models used in the simulation study of \gls{losawGD}, together with their regression function and signal features.}
    \begin{tabular}{|>{\centering\arraybackslash}p{1.7cm}|p{10cm}|>{\centering\arraybackslash}p{1.7cm}|}
    \hline
    \multicolumn{1}{|p{1.7cm}|}{Regression model} & \multicolumn{1}{p{10cm}|}{Regression function} & \multicolumn{1}{|p{1.7cm}|}{Signal features} \\
    \hline
    $8$ & $f_8(X) = \sum_{s\in \mathcal{S}} X_s$  & $\mathcal{S}$ \\
    $9$ & $f_9(X) = \sum_{s\in \mathcal{S}} \one(X_s\geq 1)$  & $\mathcal{S}$\\
    \multirow[t]{1}{*}{$10$} & 
    \parbox[t]{\linewidth}{\raggedright
        $f_{10}(X) = \one(X_1\geq 1)\cdot\one(X_3\geq 1) + \one(X_{201}\geq 1)\cdot \one(X_{204}\geq 1)$\\
        $+ \one(X_{601}\geq 1)\cdot\one(X_{602}\geq 1) \cdot \one(X_{603}\geq 1)$
    } & $\mathcal{S}$\\
    \hline
\end{tabular}
    
    \label{tab: simulation regression models NN}
\end{table}

\paragraph{\gls{NN} architecture.} The same \gls{NN} architecture is used across all simulation settings. Specifically, we implement a \gls{CNN} architecture (see \cite{lecun_gradient_1998}), consisting of two convolutional layers followed by three dense layers, resulting in a total of 4,080,001 trainable parameters. Details regarding kernel sizes and dense layer dimensions are provided in the Appendix in Figure \ref{fig:NN_architecture} (see Appendix \ref{app: NN}). The \gls{ReLU} activation function is employed in the hidden layers, while a linear activation function is used in the output layer to accommodate the continuous response. This architecture was chosen based on standard \gls{CNN} architectures, without any systematic hyperparameter tuning regarding our regression function; for standard \gls{CNN} architectures, see e.g. \cite{goodfellow_deep_2016}, and \cite{lecun_gradient_1998}.

\paragraph{Simulation and Algorithm parameters.} In each simulation, a regression function $f\in\{f_8,f_9,f_{10}\}$ is specified, and 50,000 observations are drawn from the corresponding feature and response generation models for training data, where 10,000 observations are used for testing. A \gls{CNN} with the architecture of Figure \ref{fig:NN_architecture} (see Appendix \ref{app: NN}) is then trained on the training set using both TensorFlow’s standard training loop by \cite{abadi_tensorflow_2016} and the proposed \gls{losaw} mini-batch training algorithm. In both cases, the networks are optimized using the Adam optimizer (\cite{kingma_adam_2015}) with a default learning rate of 0.001, minimizing the \gls{MSE} loss over 400 training steps (or equivalently 10 epochs) with mini-batches of size 254. Given the stochastic nature of the data generation process, each simulation is repeated 100 times using different realizations of correlation matrices of the form $\Sigma$ and independently drawn datasets $\{y^{(n)},x^{(n)}\}_{n=1,\ldots,N}$. Feature importance scores for the trained \gls{CNN}s are obtained via the saliency method by \cite{Simonyan14a} implemented as by tf-keras-vis\footnote{https://github.com/keisen/tf-keras-vis}. For the \gls{losaw} reweighting the adjustment features are selected using a correlation threshold of $0.1$ as in Remark \ref{rem: adjustment features}, but without restricting to the top $Q$ features of an initial \gls{ML} model (equivalent to setting $Q=P$).

The propensity scores are estimated using multinomial logistic regression, and inverse probability weights are adjusted using a uniformity parameter $\eta=0.2$.

\paragraph{Evaluation metrics.} In each Monte Carlo simulation run, a dataset of 10,000 observations is used to compute the test prediction $R_{test}^2$ of both CNNs. Additionally, a second test dataset, consisting of 50,000 observations, is generated where features are sampled independently from their respective marginal distributions, to compute $R_{ind}^2$ for both models, analog as in the simulation study for \gls{losawRF} (see Section \ref{sec: losaw random forest}) but here with noise ($\phi=1.0$). To evaluate the performance of both algorithms with respect to the feature importance scores, we compute the \gls{pr-AUC} as defined in Section \ref{sec: losaw random forest}. Since the \gls{pr-AUC} is nearly 1.0 across all regression functions we additionally consider the following feature importance score gap measure ($\text{FI}_{\text{gap}})$. Given $0-1$ normalized feature importance scores $s_1,\ldots,s_P$, we define
\begin{align*}
    \text{FI}_\text{gap}:= \min_{p\in \text{signal}(f)}s_p- \max_{p\not\in \text{signal}(f)}s_p.
\end{align*}
The $\text{FI}_\text{gap}$ measures both the existence of a feature importance cutoff that correctly distinguishes signal from non-signal features, as well as the margin of distinction. Therefore, it can be interpreted as a certainty of the classification. This is particularly relevant as the true number of signal features is unknown in applications and a high $\text{FI}_\text{gap}$ indicates a cutoff that can easily be determined. All stated results in Table \ref{tab: resultsNN} are the result of averaging each evaluation metric over all 100 simulation runs. 

\paragraph{Simulation results.} Our simulations show that the \gls{CNN} trained by \gls{losawGD} outperforms the \gls{CNN} trained by the standard \gls{GD} training loop, when it comes to interpretability of the feature importance scores. In particular, we observe a higher \gls{pr-AUC} and a higher $\text{FI}_\text{gap}$ for \gls{losawGD} in all simulation scenarios. For the regression function $f_{10}$, which models feature interactions in the response, we see the strongest improvement of  \gls{pr-AUC} and $\text{FI}_\text{gap}$ through \gls{losawGD} training. 
Regarding the predictability metrics, $R^2_{test}$ and $R^2_{ind}$, the simulation results show a less clear picture. For the regression function $f_{10}$, which includes interaction terms, \gls{losawGD} shows a slight improvement on predictability compared to the standard \gls{GD} training loop. For the additive regression functions $f_8$ and $f_9$ the situation is the other way around.
Overall, our simulation results show that \gls{losawGD} improves interpretability of \glspl{NN}. It may also improve generalization to new data distributions for complex regression functions, but at the cost of potential reduced predictability in other situations.

Notably, neither the \gls{CNN} architecture, the learning rate, nor the number of \gls{losawGD} training steps were subject to systematic hyperparameter tuning, and the \gls{CNN} architecture was selected without task-specific optimization. Additionally, the same training specifications were used for each simulation run even though slightly different correlation structures were used.
Hence, the observed predictive discrepancies may be attributable, at least in part, to this lack of hyperparameter and architectural refinement.  

\begin{table}
\caption{Simulation results for \gls{CNN} learned by TensorFlow and \gls{CNN} learned by \gls{losawGD} for $P=1,000$ discrete features and noise parameter $\phi=1.0$ for all three regression functions. The evaluation metrics $R^2_{ind}$, $R^2_{test}$, \gls{pr-AUC} and $\text{FI}_\text{gap}$ are average over 100 Monte Carlo runs. The algorithm that performs best in terms of the respective metric is marked in boldface.}
\label{tab: resultsNN}
\centering
\begin{tabular}{|c|l||c|c|}\hline
        regression & evaluation & \multicolumn{2}{|c|}{$P=1,000$} \\ \cline{3-4}
        function & metric & \gls{CNN} & \gls{losawGD}    \\ \hline\hline
        \multirow{4}{*}{\hfil$8$} 
        & $R^2_{test}$  & \textbf{0.451}  & 0.439   \\
        & $R^2_{ind}$  & \textbf{0.409}  &  0.399  \\       
        & pr-AUC  & 1.000  & {1.000}  \\
        & $FI_{gap}$   & 0.193 &\textbf{0.213} \\ \hline
        \multirow{4}{*}{$9$}     
        & $R^2_{test}$  & \textbf{0.452} &  0.441   \\
        & $R^2_{ind}$  &\B{0.106} &  {0.068} \\ 
        & pr-AUC  & 0.999  & \textbf{1.000}  \\
        & $FI_{gap}$        &0.293  & \textbf{0.313} \\   \hline
        \multirow{4}{*}{$10$}  
        & $R^2_{test}$  &{0.355}  & \textbf{0.366}  \\
        & $R^2_{ind}$  &0.128  & \textbf{0.138}   \\ 
        & pr-AUC  & 0.970  & \textbf{0.983}  \\
        & $FI_{gap} $       &0.009  & \textbf{0.064} \\   \hline
        
    \end{tabular}
    
\end{table}

\section{Discussion}
\label{sec: discussion}
In this paper we have introduced \gls{losaw}, a local sample-weighting approach which can be embedded into the training process of several \gls{ML} models in order to mitigate biases in feature importance scores caused by feature correlation and, more generally, feature dependence. 
We demonstrated that the \gls{losaw} approach has a natural tuning parameter, namely the effective sample size of the weighted population $\eta$, which regulates an interpretability-predictability tradeoff -- in analogy to classical tuning parameters in ML which regulate a bias-variance tradeoff.
We propose two explicit algorithms, \gls{losawRF} and \gls{losawGD}, which integrate \gls{losaw} into the training process of a \gls{RF} regressor and gradient descent mini-batch training, e.g., for \glspl{NN}. 
We conducted simulation studies that compared state-of-the-art implementations of \gls{RF} and \gls{CNN}s with their respective \gls{losaw} modifications.
We found that \gls{losaw} improves reliability of feature importance to detect signal features, at the cost of a small decrease in predictive performance on test data from the same distribution as the training data. Moreover, we found that \gls{losaw} often improves predictive behavior in the presence of a distribution shift.

In the paper, we provided details how \gls{losaw} can be integrated within \gls{RF} and \gls{NN} mini-batch training. However, we stress that it might also be integrated into the local training structure of other \gls{ML} algorithms and with other feature importance scores. 
For any other decision tree-based method one can integrate \gls{losaw} in an analog way at the local split-node selection as we have proposed here for RF. As this integration is part of the tree-training itself, it can also be combined naturally with other feature importance scores for tree-based methods. Any other parametric \gls{ML} algorithm which can be trained with mini-batch GD can also be combined with \gls{losawGD}, independent of the specific feature importance metric under consideration.
Moreover, we can also envision that \gls{losaw} can be combined with other model-agnostic feature importance metrics, independent of the training of the ML algorithms. For example, feature importance metrics which are based on mini-batches themselves, such as \gls{LOCO} mini-batch importance, see e.g., \cite{gan_model-agnostic_2023}, can be combined with \gls{losaw} via a respective weighting scheme for the aggregation of different mini-batches.
In addition, we stress that \gls{losaw} might be combined with other approaches to improve feature importance scores in the presence of feature-correlation, e.g., \cite{verdinelli_decorrelated_2024, strobl_conditional_2008}.

One limitation of \gls{losaw} might be the potential decrease in prediction accuracy for within-distribution test data. However, through its tuning parameter $\eta$ this can be controlled effectively and highlights the arguably intrinsic predictability-interpretability tradeoff. 
Another limitation of \gls{losaw} might come through the additional computational burden of calculating sample weights for each feature locally. However, we found in our simulation study that this can be mitigated by efficient pre-processing steps of selecting adjustment features. 
In future work, it will be interesting to explore further computational improvements of \gls{losaw}, for example through efficient dynamic updates of sample weights from one feature to another and with further fine-tuning how to integrate \gls{losaw} locally only in those parts of the training, where it is most effective, e.g., only at certain nodes within the tree or at certain GD steps.

\section{Acknowledgements}
We would like to thank the research group of Iris Heid for inspiring this project via applications to genomic-finemapping, as well as Gabriel Clara and Bin Yu for helpful discussions.
This project was funded by the Deutsche Forschungsgemeinschaft (DFG, German Research Foundation), project number 509149993, TRR 374.

\section{Code Availability}
Python implementations of the \gls{losawRF} and \gls{losawGD} algorithms as well as all auxiliary algorithms and simulations that are mentioned in this paper can be found at 
\url{https://git.uni-regensburg.de/behr-group-public/losaw}.

\printbibliography

\newpage
\section{Appendix}
\label{app: appendix}
\subsection{Details on split point selection algorithm}
\label{app: split point selection}
The split point selection algorithm for \gls{losawRF} is a straight-forward modification of the split point selection for standard implementations of \gls{RF} with sample weights, but we will state the algorithm in detail for completeness.

\begin{lemma}[impurity decrease formula]
\label{lem: impurity decrease formula}
    Given training labels $\{y^{(n)}\}_{n=1,\ldots,N_k}$ at a node $k$, together with a sample weight $w$ and a decomposition $[N_k]=L\sqcup R$ associated with a split 
    at the value $x$ on feature $p$, the associated weighted impurity decrease is given by the formula
    \[
    \Delta_w(x,p) = 
    \frac{T_L^2}{W_L}-\frac{(T-T_L)^2}{1-W_L}-T^2,
    \]
    where 
    \begin{align*}
        W_L \coloneqq\sum\limits_{n\in L}w_n,\quad
        T \coloneqq\sum\limits_{n\in [N_k]}w_n{y^{(n)}},\quad
        T_L \coloneqq\sum\limits_{n\in L}w_ny^{(n)}.
    \end{align*}
    Moreover, setting $S\coloneqq\sum\limits_{n\in [N_k]}w_n{y^{(n)}}^2$, 
    we get 
    $\Delta_w^{rel}(x,p)=\frac{\Delta_w}{S-T^2}$.
\end{lemma}

\begin{proof}
    We note first that $\MSE_w=S-T^2$ and $w^{-}(x,p)=W$ by Definition 
    \ref{def: rel imp dec}.
    Furthermore, we may compute the term $w^{-}(x,p)\cdot\MSE_w(x,p)^{-}$ as follows.
    \begin{align*}
        &w^{-}(x,p)\cdot\MSE_w(x,p)^{-} 
        \\
        &=
        w^{-}(x,p)\cdot\left(\frac{1}{w^{-}(x,p)}\sum\limits_{n=1}^{N_k}\one(x_p^{(n)}\leq x)\cdot w_n{y_p^{(n)}}^2-
        \left(\frac{1}{w^{-}(x,p)\cdot}\sum\limits_{n=1}^{N_k}\one(x_p^{(n)}\leq x)\cdot w_n{y_p^{(n)}}\right)^2\right)
        \\
        &=
        W\cdot\left(\frac{1}{W}\sum\limits_{n\in L}w_n{y_p^{(n)}}^2-\frac{1}{W^2}\left(\sum\limits_{n\in L}w_ny^{(n)}\right)^2\right)
        =
        \sum\limits_{n\in L}w_n{y_p^{(n)}}^2-
        \frac{T_L^2}{W}.
    \end{align*}
    Noting that 
    $T-T_L = \sum\limits_{n\in [N_k]}w_ny_p^{(n)}-\sum\limits_{n\in L}w_ny_p^{(n)}=\sum\limits_{n\in R}w_ny_p^{(n)}$, we get analogously that
    \begin{align*}
        w^{+}(x,p)\cdot\MSE_w(x,p)^{+} 
        &=
        \sum\limits_{n\in R}w_n{y_p^{(n)}}^2-\frac{(T-T_L)^2}{1-W}
    \end{align*}
    From this it follows that
    \begin{align*}
        \Delta_w(x,p) &=
        \MSE_w - w^{-}(x,p)\cdot\MSE_w(x,p)^{-}-
        w^{+}(x,p)\cdot\MSE_w(x,p)^{+}
        \\
        &=
        S-T^2-\left(\sum\limits_{n\in L}w_n{y_p^{(n)}}^2-\frac{T_L^2}{W}\right)-
        \left(\sum\limits_{n\in R}w_n{y_p^{(n)}}^2-\frac{(T-T_L)^2}{1-W}\right)
        \\
        &=
        \frac{T_L^2}{W}+\frac{(T-T_L)^2}{1-W}-T^2+S-
        \sum\limits_{n\in [N_k]}w_n{y_p^{(n)}}^2
        =
        \frac{T_L^2}{W}+\frac{(T-T_L)^2}{1-W}-T^2.
    \end{align*}
\end{proof}

\begin{Algorithm}[split point selection for continuous features]
\label{alg: split point selection for continuous features}
\begin{algorithm}

\begin{algorithmic}\\
    \State{\textbf{Input:}}
    training dataset $\{(y^{(n)},x^{(n)})\}_{n=1,\ldots,N_k}$ at a node $k$, sample weight $w$, a splitting feature $p\in [P]$ such that 
    $(x_p^{(1)},\ldots,x_p^{(N_k)})$ is in increasing order.
    \State{\textbf{Output:}}
    an optimal split point $x=\argmax\limits_{x\in\{x_p^{(1)},\ldots,x_p^{(N_k)}\}}\Delta_w^{rel}(x,p)$ and 
    associated relative weighted impurity decrease
    $\Delta_w^{rel}(x,p)$.
    \State $S\gets\sum\limits_{n=1}^{N_k}w_n{y^{(n)}}^2,\quad T\gets\sum\limits_{n=1}^{N_k}w_ny^{(n)}$
    \State $W(0)\gets 0,\quad T(0)\gets 0$
    \Comment{Initialize recursion}
    \State $x\gets\emptyset,\quad\Delta\gets 0$
    \Comment{Initialize split}
    \For{$n=1,\ldots,N_k$}
    \State $W(n)\gets W(n-1)+w_n$
    \State $T(n)\gets T(n-1)+w_ny^{(n)}$
    \If{$W(n)\in \{0,1\}$}
    \State $\Delta_{temp}\gets 0$
    \Else
    \State $\Delta_{temp}\gets \frac{T(n)^2}{W(n)}+\frac{(T-T(n))^2}{1-W(n)}-T^2$
    \If{$\Delta_{temp}>\Delta$}
    \State $x\gets x_p^{(n)},\quad\Delta\gets\Delta_{temp}$
    \Comment{Update split data}
    \EndIf
    \EndIf
    \EndFor
    \State{\textbf{Return:}}
    $(x,\frac{\Delta}{S-T^2})$
\end{algorithmic}
\end{algorithm}
\end{Algorithm}

\begin{Algorithm}[split point selection for discrete features]
\label{alg: split point selection for discrete features}
\begin{algorithm}

    \begin{algorithmic}\\
        \State\textbf{Input:} 
        training dataset $\{(y^{(n)}, x^{(n)})\}_{n=1,\ldots,N_k}$ at a node $k$,
        sample weight $w$, 
        a discrete feature $p\in[P]$,
        the possible values $\{\mathbf{x}_1,\ldots,\mathbf{x}_K\}$ of feature $p$.
        \State\textbf{Output:}
        the optimal splitting point $x$ and corresponding 
        relative weighted impurity decrease $\Delta_w(x,p)$.
        \State $S\gets\sum\limits_{n=1}^{N_k}w_n{y^{(n)}}^2,\quad T\gets\sum\limits_{n=1}^{N_k}w_ny^{(n)}$
        \State{$(x\gets\emptyset, \Delta\gets 0)$}
        \For{$k=1,\ldots,K$}
        \State $L\gets\{n\in [N_k]\mid x_p^{(n)}\leq\mathbf{x}_k\}$
        \State $W\gets\sum\limits_{n\in L}w_n$
        \If{$W\not\in\{0,1\}$}
        \State $T_{L}\gets \sum\limits_{n\in L}w_ny^{(n)}$
        \State $\Delta_{temp}\gets\frac{T_L^2}{W}-\frac{(T-T_L)^2}{1-W}-T^2$
        \If{$\Delta_{temp}>\Delta$}
        \State $x\gets\textbf{x}_k,\quad\Delta\gets\Delta_{temp}$
        \EndIf
        \EndIf
        \EndFor
        \State{\textbf{Return:}} 
        $(x,\frac{\Delta}{S-T^2})$
    \end{algorithmic}
\end{algorithm}
\end{Algorithm}

\begin{lemma}
\label{lem: complexities of split point selection}
    \begin{enumerate}
        \item 
        The split point $x$ computed by Algorithm 
        $\ref{alg: split point selection for continuous features}$ satisfies 
        \[
        x= \argmax\limits_{x\in\{x_p^{(1)},\ldots,x_p^{(N_k)}\}}\Delta_w^{rel}(x,p).
        \]
        Moreover, combined with a sorting algorithm ensuring that the training dataset \\
        $(y^{(n)},x^{(n)})_{n=1,\ldots,N_k}$ 
        and the sample weight $w$ are sorted in increasing order of $x_p^{(n)}$, it has a runtime complexity of 
        $O(N_k\log N_k)$.
        \item 
        The split point $x$ computed by Algorithm 
        \ref{alg: split point selection for discrete features} satisfies 
         \[
        x= \argmax\limits_{x\in\{\textbf{x}_1,\ldots,\textbf{x}_K\}}\Delta_w^{rel}(x,p).
        \]
        Moreover, the runtime complexity of the algorithm is given by 
        $O(KN_k)$. 
        \end{enumerate}
\end{lemma}

\begin{proof}
    For both parts, it follows directly from Lemma \ref{lem: impurity decrease formula} that the value $\Delta_{temp}$ computed in the respective for-loop is equal to
    $\Delta_w(x_p^{(n)},p)$ and 
    $\Delta_w(\mathbf{x}_k,p)$, respectively.
    As both algorithms update the split datum whenever the current weighted impurity decrease is higher, 
    the pair $(x,\Delta)$ after termination of the for-loop corresponds to the split with the largest weighted impurity decrease.
    It remains to show the assertions on runtime complexity.
    \begin{enumerate}
        \item The updates within the for-loop are of $O(1)$, so in combination with the computation of $S$ and $T$ in the beginning, the entire algorithm has a complexity of $O(N_k)$. 
        In combination with a sorting algorithm (which is of $O(N_k\log N_k)$ complexity), we can compute the optimal splitting point in a $O(N_k\log N_k)$ time.
        \item 
        The sums $S$ and $T$ can be computed in $O(N_k)$ time by iterating over the sample. Likewise, in the for-loop, the set $L$ and left child weight $W$ can be computed in $O(N_k)$ time, whereas the remaining steps are of constant complexity. As this loop consists of $K$ iterations, we end up with a complexity of $O(KN_k)$. 
    \end{enumerate}
    
\end{proof}

\subsection{Details on adjusting weights for effective sample size}
\label{app: effective sample size}
As mentioned in Section \ref{sec: interpretation-prediction-tradeoff}, the minimal relative effective sample size is an important tuning parameter for our \gls{losaw} approach that mediates the tradeoff between decorrelation (interpretation) and effective sample size (prediction). 
We will now define the algorithm which we use to enforce a given effective sample size bound upon a sample weight.

\begin{Algorithm}
\label{alg: enforce maximal weight threshold}

    \begin{algorithm}
    \begin{algorithmic}\\
    \State\textbf{Inputs:} normalized sample weight $w\in\R_{\geq 0}^N$, minimal weight threshold $\theta\in[\frac{1}{N},1]$
    \State\textbf{Output:} a modified normalized sample weight $w_{\theta}$ such that $\max(w_\theta)\geq\theta$.
        \State $i\coloneqq 0$, $w^{(0)}\coloneqq w$
        \While{$\max(w^{(i)})>\theta$}
        \State $S^{(i)}\coloneqq\{n\in[N]\mid w^{(i)}_n\geq\theta\}$
        \State $\text{excess}^{(i)}\coloneqq\sum\limits_{n\in S^{(i)}}(w_n^{(i)}-\theta)$
        \Comment{compute excess weight}
        \For{$n\in[N]$}
        \If{$n\in S^{(i)}$}
        \State $w_n^{(i+1)}\gets\theta$
        \Comment{cut off weights at threshold}
        \Else
        \State $w_n^{(i+1)}\gets w_n^{(i)}+\frac{\text{excess}}{N-\vert S^{(i)}\vert}$
        \Comment{distribute excess weight}
        \EndIf
        \EndFor
        \State $i\gets i+1$
        \EndWhile
        \State{\textbf{return} }$w_\theta\coloneqq w^{(i)}$
    \end{algorithmic}
    \end{algorithm}
\end{Algorithm}

\begin{proposition}
\label{prop: enforce maximal weight threshold}
    Given a normalized sample weight $w\in\R_{\geq 0}^N$ and a real threshold $\theta\in[\frac{1}{N},1]$, the Algorithm \ref{alg: enforce maximal weight threshold} converges and defines a well-defined normalized sample weight.
\end{proposition}

\begin{proof}
    First, note that, clearly, after each run of the inner for-loop $w$ remains a well-defined normalized sample weight, with non-negative entries.
    Second, notice that for each $n\in S^{(i)}$, the corresponding sample weight is set to $\theta$ after the for-loop, and in particular we have $n\in S^{(i+1)}$. 
    So the $S^{(i)}$ are subsets of the finite set $[N]$ and ordered by inclusion. Therefore, after at most $N$ steps, they must become stable, such that there exists some $i\in[N]$ with $S^{(i)}=S^{(i+1)}$. But then, by the previous argument we have $w^{(i+1)}_n=\theta$ for each 
    $n\in S^{(i+1)}$ and the while-loop halts. This proves that the Algorithm \ref{alg: enforce maximal weight threshold} terminates after at most $N$ steps and returns a well-defined normalized sample weight.
\end{proof}

\begin{Algorithm}[binary search for sample weight threshold]
\label{alg: binary search for weight}

Using the following binary search algorithm, we can enforce a given relative effective sample weight $\eta\in (0,1]$ up to a given tolerance $\alpha$ by redistributing the sample weight
\begin{algorithm}
    \begin{algorithmic}\\
        \State\textbf{Inputs: }
        normalized sample weight $w\in\R_{\geq 0}^N$, 
        minimal relative effective sample size $\eta\in (0,1]$,
        tolerance $\alpha\in [0,1]$
        \State\textbf{Output: }
        a modified normalized sample weight $w'\in\R_{\geq 0}^N$ such that 
        $s_{rel}(w')\in (\eta-\alpha,\eta+\alpha)$
        \State$\theta_{\min}\gets\frac{1}{N\eta}$
        \State$\theta_{\max}\gets 1$
        \State$w'\gets w$
        \While{$\vert s_{rel}(w')-\eta\vert>\alpha$}
        \State$\theta\gets(\theta_{\min}+\theta_{\max})/2$
        \If{$s_{rel}(w')\geq\eta$}
        \State$\theta_{\min}\gets\theta$
        \Comment{search in upper half of interval}
        \Else
        \State$\theta_{\max}\gets\theta$
        \Comment{search in lower half of interval}
        \EndIf
        \State$w'\gets w_\theta$
        \Comment{As computed by the Algorithm \ref{alg: enforce maximal weight threshold}}
        \EndWhile
        \State\textbf{return }$w'$
    \end{algorithmic}
\end{algorithm}
\end{Algorithm}

\begin{remark}
\label{rem: assumptions for binary search depth, weight iterations}
    In the proof of Proposition \ref{prop: enforce maximal weight threshold} we show that the algorithm there converges after at most $N$ steps. 
    However, in practice we found that for $\theta$ bounded away from the minimum $\frac{1}{N}$ (corresponding to uniform weight), the number of iterations remains relatively constant. 
    More precisely, for $\theta\geq\frac{3}{2N}$, in our simulations, the Algorithm always converged within at most $4$ iterations.
    Likewise, we observed the depth in the binary search algorithm 
    \ref{alg: binary search for weight} to be quite consistent.
    We therefore assume in the following that these weight modification algorithms require a constant number of iterations and a constant binary search depth, such that they can be performed in $O(N)$ time, as long as $\eta$ is bounded away from $1$. 
    For $\eta=1$ all sample weights are uniform and 
    \gls{losawRF} is identical to \gls{RF} but requires higher computational cost to include the estimation of propensities and the weight modification algorithms. 
    We therefore advice to restrict to choose $\eta$ within the interval $(0,\frac{2}{3}]$ for applications of \gls{losaw}.
\end{remark}

\subsection{Details on computational complexity of losawRF}
\label{app: computational complexity}

We begin by computing the complexity class associated to computing the split at a single node in the \gls{losawRF}.

\begin{lemma}[complexity class at a node sample]
\label{lem: complexity class at a node sample}
    We consider a training sample at node $k$ consisting of $P$ features and $N_k$ observations and assume that $m_{try}=O(P)$ features are considered for the node split.
    We assume moreover that $Q$ features are used for the estimation of propensity scores, and that a sample weight can be modified within $O(N_k)$ time to ensure a lower effective sample size bound $\eta$, as stated in Remark \ref{rem: assumptions for binary search depth, weight iterations}.
    Under these assumptions, the computational complexity of a \gls{losawRF} node to compute the optimal splitting feature and split point is given by
    \begin{enumerate}
        \item $O(PQN_kK^2)$ for discrete features
        \item $O(PQN_k + PN_k\log N_k)$ for continuous features.
    \end{enumerate}
\end{lemma}

\begin{proof}
    \begin{enumerate}
        \item 
        For the estimation of propensity scores with discrete features, the $Q$ adjustment features are embedded into a $KQ$-dimensional space using one-hot encoding, for which then a gradient-based $K$-class logistic regression model is fitted requiring complexity $O(QN_kK^2)$ 
        \cite{bishop_pattern_2006}, and likewise for the prediction of class probabilities from the fitted model. 
        The computation of stabilization scores using frequency count is of mere $O(KN_k)$ complexity and therefore vanishes against the remaining terms.
        Given a splitting feature, the computation of the optimal splitting point can be done by comparing the $K$ possible split values, computing for each the relative weighted impurity decrease in $O(N_k)$ time, which has a total complexity of $O(KN_k)$. 
        Since the propensity estimation, weight computation and split point selection have to be performed for each available splitting feature, the entire 
        split selection algorithm on discrete features has complexity
        $O(m_{try}(N_k + QN_kK^2 + KN_k)) =
        O(PQN_kK^2)$.
        \item 
        For the estimation of propensity scores with continuous features, a linear regression model has to be fitted for each available splitting feature. 
        Each of these linear models requires the covariance matrix of the $Q$ adjustment features, so it suffices to compute this matrix a single time in $O(Q^2N_k)$ complexity. 
        For each splitting feature, we additionally have to compute its covariance with the adjustment features 
        ($O(QN_k)$) and fit a linear regression model, which has $O(Q^2)$ complexity using conjugate gradient \cite{bottou_optimization_2018}.
        Computing the predictions from this linear model is done in $O(QN_k)$ time, and estimating the variance of the residual error, as well as computing the respective propensity scores from it is $O(N_k)$.
        Also the estimation of stabilization scores from a normal distribution is of $O(N_k)$ complexity. 
        As shown in Lemma \ref{lem: complexities of split point selection}, the split point selection Algorithm \ref{alg: split point selection for continuous features} has another $O(N_k\log N_k)$ complexity, leading to a total of 
        $O(Q^2N_k+m_{try}\cdot(QN_k+Q^2+QN_k+N_k+N_k+N_k\log N_k)) = 
        O(PN_k(Q + \log N_k))$.
    \end{enumerate}
\end{proof}

\paragraph*{Proof of Proposition \ref{prop: time complexity classes for losawRF}}
\quad 

\begin{proof}
    It suffices to compute the complexity of a single tree estimator. 
    We note that at depth $d$, each node sample can have at most $(1-\gamma)^dN$ observations. In particular, the maximal depth $D$ of the tree must satisfy
    \[
    (1-\gamma)^DN\geq 1\Leftrightarrow D\leq-\log_{1-\gamma}N=\log_{\frac{1}{1-\gamma}}N.
    \]
    Let $I$ be an indexing of all of the nodes of the tree, and for $i\in I$, let $N_i$ denote the corresponding node sample size and $\delta(i)$ denote the depth of the node.
    For a given depth $d$, we observe that the samples for all nodes at that depth are disjoint an hence 
    \[\sum\limits_{i\in I,\delta(i)=d}N_i\leq N\]
    For the discrete features, we may therefore use the formula for the node complexity from Lemma \ref{lem: complexity class at a node sample} to estimate the complexity of the tree as
    \begin{align*}
        \sum\limits_{i\in I}O(PQN_iK^2) 
        &=
        \sum\limits_{d=0}^{-\log_{1-\gamma}N}\sum\limits_{i\in I, \delta(i)=d}O(PQN_iK^2)
        =
        O\left(PQK^2\sum\limits_{d=0}^{-\log_{1-\gamma}N}\sum\limits_{i\in I, \delta(i)=d}N_i\right)\\
        &\leq
        O(PQK^2\sum\limits_{d=0}^{-\log_{1-\gamma}N}N) =
        O(PQK^2N\log_{\frac{1}{1-\gamma}}N).
    \end{align*}
    For the discrete features, we derived the complexity class $O(PQN_k + PN_K\log N_k)$ for a single node $k$ in Lemma \ref{lem: complexity class at a node sample}.
    With the first term, we can repeat the above argument and get the complexity 
    $O(PQN\log_{\frac{1}{1-\gamma}}N)$. To estimate also the second term, we note that the function $f(x)=x\cdot\log(x)$ with the continuation $f(0)\coloneqq 0$ is convex on $\mathbb{R}_{\geq 0}$, and therefore we find that for any $x>0$, $\lambda\in [0,1]$, 
    \[
    \lambda f(0)+(1-\lambda)f(x)\geq f(\lambda\cdot 0+(1-\lambda)x),\quad
    (1-\lambda) f(0)+\lambda f(x)\geq f((1-\lambda) \cdot 0+\lambda x)
    \]
    and so 
    $f(x)\geq f(\lambda x)+f((1-\lambda)x).$
    In particular, for any node $i\in I$, where its child nodes contain a proportion of $\lambda$ and $1-\lambda$ of its sample size, we find that 
    $f(N_i)\geq f(\lambda N_i)+f((1-\lambda)N_i)$. 
    With this, it follows that at any depth $d>0$,
    \[
    \sum\limits_{i\in I, \delta(i)=d}f(N_i)\leq 
    \sum\limits_{i\in I,\delta(i)=d-1}f(N_i)
    \]
    by estimating the terms of two child nodes against their parent node. 
    Recursively, each of these sums is therefore bounded from above by $f(N)$.
    \begin{align*}
        \sum\limits_{i\in I}O(PN_i\log N_i) &=
        \sum\limits_{d=0}^{-\log_{1-\gamma}N}\sum\limits_{i\in I,\delta(i)=d}O(PN_i\log N_i) =
        O\left(P\sum\limits_{d=0}^{-\log_{1-\gamma}N}\sum\limits_{i\in I,\delta(i)=d}f(N_i)\right)
        \\
        &\leq
        O\left(P\sum\limits_{d=0}^{-\log_{1-\gamma}N}f(N)\right)
        =
        O(PN\log N\log_{\frac{1}{1-\gamma}}N).
    \end{align*}
    By combining both terms, we get the complexity class as stated in the Proposition.
\end{proof}

\subsection{Details on discrete feature generation in simulation study}
\label{app: discrete distribution with numerical optimization}
For the data generation model with discrete features in our simulation studies we used a joint distribution $\P_{joint}$, where the marginal distributions of the features are given by $\P_1,\ldots,\P_P$ with mean values $\mu_1,\ldots,\mu_P$ and standard deviations $\sigma_1,\ldots,\sigma_P$, and the correlation matrix associated to $\P_\Sigma$ minimizes the Euclidean distance to a given correlation matrix $\Sigma$.
We will discuss how $\P_{joint}$ can be found by solving a convex optimization problem. 
Afterwards we will specify the specific choices for the marginals which we used in our simulation studies.
Denoting by $X_1,\ldots,X_P$ the random variables with distribution $\P_{joint}$ and by $\Omega_1,\ldots,\Omega_P$ their sets of values, we know that $\P_{joint}$ solves the constraint optimization problem

\begin{align*}
    \P_{joint} &=
    \argmin_{\P\in\mathcal{D}(\Omega_1\times\ldots\times\Omega_P)}\Vert \corr(\P)-\Sigma\Vert_2^2 
    \\
    &\text{s.t.}\quad
    \forall p\in\{1,\ldots,P\}\ \forall x\in\Omega_p\colon\P(X_p=x)=\P_p(x),
\end{align*}
where $\mathcal{D}(\Omega_1\times\ldots\times\Omega_P)$ denotes the set of probability distributions on 
$\Omega_1\times\ldots\times\Omega_P$. 
In order to rephrase it as a convex optimization problem, we note that each distribution $\P\in\mathcal{D}(\Omega_1\times\ldots\times\Omega_P)$ can be identified with its probability vector and hence, 
\begin{align*}
    \P_{joint} &=
    \argmin\limits_{\P\in\R^{\Omega_1\times\ldots\times\Omega_P}}
    \left\Vert\left[\frac{1}{\sigma_p\sigma_q}\sum\limits_{x_1\in\Omega_1}\ldots\sum\limits_{x_P\in\Omega_P}
    \P_{x_1,\ldots,x_P}(x_p-\mu_p)(x_q-\mu_q)\right]_{p,q}-\Sigma\right\Vert_2^2
    \\
    &\text{s.t.}\quad
\forall p\in\{1,\ldots,P\}\forall x\in\Omega_p\colon
\sum\limits_{x_1\in\Omega_1}\ldots\widehat{\sum\limits_{x_p\in\Omega_p}}\ldots\sum\limits_{x_P\in\Omega_P}
\P_{x_1,\ldots,x,\ldots,x_P}=\P_p(x),
    \\
    &
    \sum\limits_{x_1\in\Omega_1}\ldots\sum\limits_{x_P\in\Omega_P}\P_{x_1,\ldots,x_P}=1,
    \\
    &
    \forall x_1,\ldots,x_P\in\Omega_1\times\ldots\times\Omega_P\colon\P_{x_1,\ldots,x_P}\geq 0.
\end{align*}
Here the operator ``$\widehat{\quad}$'' means that the $p$-th summation operator is suppressed, i.e, the sum is only taken over the other indexing sets. 
Although the convexity allows for an efficient computation of the distribution $\P_{joint}$, the dimension of this optimization problem grows as $\Omega_1\times\ldots\times\Omega_P$ with the number of variables (i.e. exponential in $P$ under the assumption that $\vert\Omega_p\vert$ remains constant). In particular in combination with the normality constraint, this leads to a fast decay of the entries of $\P_{joint}$ and hence numerical instability with growing dimension. 
We therefore restricted to only compute the joint distribution for a block of $P=6$ variables for the \gls{losawRF} simulation study, and for blocks of $P=5$ variables for the \gls{losawGD} simulation study.

For the marginal distribution $\P_p$ of each discrete feature we used a centered binomial distribution with $2$ trials and success probability $0.5$. More precisely, each feature can take values in $\{-1,0,1\}$ with probabilities
\[
\P_p(x) = 
\begin{cases}
    0.25 & \text{if } x =\pm 1\\
    0.5 & \text{if } x= 0
\end{cases}
\]
and has mean $\mu_p=0$ and variance $\sigma_p^2=0.5$.

Recall that for the \gls{losawRF} simulation study a fixed correlation matrix $\Sigma$ is considered. In contrast, for the \gls{losawGD} simulation study $\Sigma$ was also generated randomly.
More precisely, in each Monte Carlo simulation run a block diagonal correlation matrix $\Sigma$ is generated out of one randomly generated sub-block matrix $\sigma$ in the following way. We construct positive definite correlation matrices of size $5 \times 5$ that contain both highly and weakly correlated feature pairs. The procedure begins by drawing a random matrix and forming its Gram matrix to ensure positive semi-definiteness and symmetry. This matrix is then standardized along the diagonal to unit variance, resulting in a valid correlation matrix. To impose structured dependencies among specific feature pairs, selected off-diagonal entries are overwritten with $\alpha$ sampled uniformly from a high correlation interval $(0.7, 1)$, while $\beta$ is sampled from a lower interval $(0, 0.4)$. The process is repeated with varying seeds until the resulting matrix satisfies the conditions of positive definiteness and exact symmetry, ensuring that each sampled block $\sigma$ is statistically admissible for modeling a realistic joint feature distribution.

\subsection{Details on NN architecture used in losawGD simulation study}
\label{app: NN}
Figure \ref{fig:NN_architecture} illustrates the exact configuration of the \gls{CNN} architecture employed in all simulation scenarios of Section \ref{sec: losawNN simulation study}. The input to the network is tabular data with 3000 features, represented as a 2D tensor of shape $(1000,1,1)$. It passes through two 1D convolutional layers with 32 and 64 filters (kernel sizes 5 and 3, respectively), each followed by \gls{ReLU} activation. The output is flattened and processed by three dense layers with 64, 32, and 1 unit(s), respectively. \gls{ReLU} is used in the hidden layers, while a linear activation is applied in the output to suit the regression task. 
\usetikzlibrary{positioning}
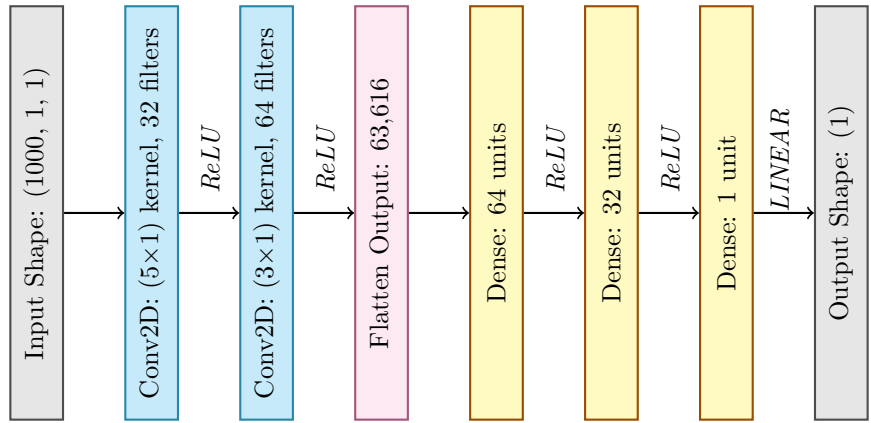
\begin{figure}
    \centering
\begin{tikzpicture}[
    layer/.style={rectangle, draw=black, minimum height=0.7cm, minimum width=5.5cm, font=\small, align=center, rotate=90},
    conv/.style={layer, draw=cyan!60!black, fill=cyan!20},
    dense/.style={layer, draw=orange!60!black, fill=yellow!30},
    flat/.style={layer, draw=magenta!60!black, fill=magenta!10},
    io/.style={layer, draw=gray!60!black, fill=gray!20},
    ->, thick,
    node distance=0cm and 0cm,
]

\node[io] (input) {Input Shape: (1000, 1, 1)};
\node[conv, below=of input.south, yshift=-0.8cm] (conv1) {Conv2D: (5×1) kernel, 32 filters};
\node[conv, below=of conv1, yshift=-0.8cm] (conv2) {Conv2D: (3×1) kernel, 64 filters};
\node[flat, below=of conv2, yshift=-0.8cm] (flatten) {Flatten Output: 63,616};
\node[dense, below=of flatten, yshift=-0.8cm] (dense1) {Dense: 64 units};
\node[dense, below=of dense1, yshift=-0.8cm] (dense2) {Dense: 32 units};
\node[dense, below=of dense2, yshift=-0.8cm] (dense3) {Dense: 1 unit};
\node[io, below=of dense3, yshift=-0.8cm] (output) {Output Shape: (1)};

\draw[->] (input.south) -- node[above, font=\small\itshape] {} (conv1.north);
\draw[->] (conv1.south) -- node[above, font=\small\itshape,rotate=90, xshift=0.7cm, yshift=-8] {ReLU} (conv2.north);
\draw[->] (conv2.south) -- node[above, font=\small\itshape,rotate=90, xshift=0.7cm, yshift=-8] {ReLU} (flatten.north);
\draw[->] (flatten.south) -- node[above, font=\small\itshape] {} (dense1.north);
\draw[->] (dense1.south) -- node[above, font=\small\itshape, rotate=90, xshift=0.7cm, yshift=-8] {ReLU} (dense2.north);
\draw[->] (dense2.south) -- node[above, font=\small\itshape, rotate=90, xshift=0.7cm, yshift=-8] {ReLU} (dense3.north);
\draw[->] (dense3.south) -- node[above, font=\small\itshape, rotate=90, xshift=0.7cm, yshift=-6] {LINEAR} (output.north);

\end{tikzpicture}
    \caption{Architecture of the \gls{NN} model used in this study in both training algorithms. The model processes input data of shape (1000, 1, 1) through a series of convolutional layers with \gls{ReLU} activation functions, followed by flattening and three fully connected (dense) layers. The final output is a scalar prediction. Each layer's configuration, including kernel size, number of filters or units, and activation function, is indicated.}
    \label{fig:NN_architecture}
\end{figure}

\subsection{Detailed results of the losawRF simulation study}
\label{app: details on simulation study}
In Section \ref{sec: losaw RF simulation study} we presented the results of our simulation study for the case of $P=100$ features and low noise parameter $\phi=0.1$ as the results remain quite consistent with variation in these parameters. 
The results of all other parameter combinations can be found in Tables \ref{tab: 10, discrete}-\ref{tab: 100, continuous}.

\label{app: all results}
\begin{table}[ht]
    \centering
    \caption{Same as Table \ref{tab: results discrete}, but for $P=10$ discrete features  and for both configurations of the noise parameter $\phi$.}
    \begin{tabular}{|c|l||c|c|c|c|c|c|c|c|}\hline
        regression & \centering valuation &
        \multicolumn{4}{|c|}{$\phi = 0.1$} & \multicolumn{4}{|c|}{$\phi = 1.0$} \\ \cline{3-10}
        function & \centering metric & 
        \multicolumn{2}{|c|}{$N = 500$} & \multicolumn{2}{|c|}{$N=5000$} &
        \multicolumn{2}{|c|}{$N = 500$} & \multicolumn{2}{|c|}{$N=5000$} 
        \\ \cline{3-10}
        &   & RF    & losaw & RF    & losaw & RF    & losaw & RF    & losaw \\ 
        &   &       & RF    &       & RF    &       & RF    &       & RF \\ \hline\hline
        \multirow{3}{*}{$1$}  
        & $R^2_{test}$ & 0.870 & \B{0.878}  & 0.903 & \B{0.905}  & \B{0.465} & 0.462  & \B{0.504} & 0.504 \\
        & $R^2_{ind}$ & 0.590 & \U{\B{0.685}}  & 0.918 & \B{0.938} & 0.615 & \B{0.621}  & \B{0.912} & 0.911 \\
        & pr-AUC & 1.000 & 1.000 & 1.000 & 1.000 & 0.997 & \B{1.000} & 1.000 & 1.000 \\   \hline
        \multirow{3}{*}{$2$}  
        & $R^2_{test}$ & 0.864 & \B{0.872}  & 0.891 & \B{0.892}  & \B{0.474} & \B{0.470}  & \B{0.475} & 0.475 \\
        & $R^2_{ind}$ & 0.705 & \B{0.750}  & 0.876 & \B{0.905} & \B{0.651} & {0.649}  & 0.869 & \B{0.872} \\
        & pr-AUC & 0.999 & \B{1.000} & 1.000  & 1.000   & \B{0.998} & 0.987 & 1.000  & 1.000 \\   \hline
        \multirow{3}{*}{$3$}  
        & $R^2_{test}$ & 0.883 & \B{0.889} & 0.904 & \B{0.907}  & \B{0.459} & 0.456 & 0.506 & \B{0.507} \\
        & $R^2_{ind}$ & 0.466 & \U{\B{0.629}} & 0.647 & \U{\B{0.804}}  & 0.392 & \U{\B{0.487}} & 0.634 & \U{\B{0.735}} \\
        & pr-AUC & 0.417 & \U{\B{0.976}} & 0.417 & \U{\B{1.000}} & 0.433 & \U{\B{0.725}} & 0.417 & \U{\B{0.986}} \\   \hline
        \multirow{3}{*}{$4$}  
        & $R^2_{test}$ & 0.878 & \B{0.882}  & 0.900 & \B{0.903}  & \B{0.468} & 0.466 & 0.498 & \B{0.499} \\
        & $R^2_{ind}$ & 0.439 & \U{\B{0.532}}  & 0.680 & \U{\B{0.828}}  & 0.508 & \U{\B{0.586}}  & 0.652	 & \B{0.756} \\
        & pr-AUC & 0.516 & \U{\B{0.803}} & 0.515 & \U{\B{0.929}} & 0.558 & \U{\B{0.873}} & 0.517 & \U{\B{0.922}} \\   \hline
        \multirow{3}{*}{$5$}  
        & $R^2_{test}$ & 0.882 & \B{0.888}  & 0.902 & \B{0.904}  & \B{0.473} & 0.471  & 0.482 & \B{0.484}\\
        & $R^2_{ind}$ & 0.787 & \U{\B{0.854}}  & 0.895 & \B{0.940}  & 0.693 & \U{\B{0.744}} & 0.857 & \B{0.896}  \\
        & pr-AUC & 0.948 & \U{\B{0.999}}  & 0.904 & \U{\B{1.000}} & 0.810 & \U{\B{0.928}}  & 0.993 & \B{1.000} \\   \hline
        \multirow{3}{*}{$6$}  
        & $R^2_{test}$ & 0.881 & \B{0.888} & 0.904 & \B{0.905}  & \B{0.461} & {0.461} & 0.482 & \B{0.482} \\
        & $R^2_{ind}$ & 0.781 & \U{\B{0.854}} & 0.943 & \B{0.958}  & 0.738 & \B{0.771} & 0.928 & \B{0.936}\\
        & pr-AUC & 1.000 & 1.000 & 1.000 & 1.000  & 0.993 & \B{0.998}  & 1.000 & 1.000 \\   \hline 
        \multirow{3}{*}{$7$}  
        & $R^2_{test}$ & 0.879 & \B{0.888} & 0.906 & \B{0.907}  & \B{0.470} & 0.466  & 0.486 & \B{0.486} \\
        & $R^2_{ind}$ & 0.662 & \U{\B{0.756}} & 0.874 & \U{\B{0.928}} & 0.631 & \B{0.674} & 0.844 & \B{0.882} \\
        & pr-AUC & 0.706 & \U{\B{0.936}}  & 0.818 & \U{\B{0.993}}  & 0.696 & \U{\B{0.871}}  & 0.820 & \U{\B{0.976}} \\   \hline
    \end{tabular}
    \label{tab: 10, discrete}
\end{table}

\begin{table}[ht]
    \centering
    \caption{Same as Table \ref{tab: results continuous}, but for $P=10$ continuous features  and for both configurations of the noise parameter $\phi$.}
    \begin{tabular}{|c|l||c|c|c|c|c|c|c|c|}\hline
        regression & evaluation &
        \multicolumn{4}{|c|}{$\phi = 0.1$} & \multicolumn{4}{|c|}{$\phi = 1.0$} \\ \cline{3-10}
        function & metric & 
        \multicolumn{2}{|c|}{$N = 500$} & \multicolumn{2}{|c|}{$N=5000$} &
        \multicolumn{2}{|c|}{$N = 500$} & \multicolumn{2}{|c|}{$N=5000$} 
        \\ \cline{3-10}
        &   & RF    & losaw & RF    & losaw & RF    & losaw & RF    & losaw \\ 
        &   &       & RF    &       & RF    &       & RF    &       & RF \\ \hline\hline
        \multirow{3}{*}{\hfil$1$}  
        & $R^2_{test}$ & 0.873 & \B{0.874}  & 0.898 & \B{0.989} & \B{0.462}  & 0.457  & 0.487 & \B{0.488} \\
        & $R^2_{ind}$ & 0.642 & \B{0.678}  & 0.769 & \B{0.783} & 0.599 & \B{0.616}  & 0.759 & \B{0.772} \\
        & pr-AUC & 1.000 & 1.000 & 1.000 & 1.000 & 0.997 & \B{1.000} & 1.000 & 1.000  \\   \hline
        \multirow{3}{*}{$2$}  
        & $R^2_{test}$ & \B{0.848} & 0.846  & \B{0.889} & 0.887 & \B{0.442} & 0.435  & \B{0.483} & 0.479\\
        & $R^2_{ind}$ & 0.667 & \B{0.695}  & 0.797 & \B{0.820} & \B{0.611} & 0.610  & 0.782 & \B{0.788}\\
        & pr-AUC & 1.000 & 1.000 & 1.000  & 1.000 & \B{0.976} & 0.960 & 1.000  & 1.000 \\   \hline
        \multirow{3}{*}{$3$}  
        & $R^2_{test}$ & \B{0.862} & 0.859 & 0.891 & \B{0.891} & \B{0.460} & 0.452 & \B{0.484} & 0.483 \\
        & $R^2_{ind}$ & 0.419 & \U{\B{0.530}} & 0.520 & \U{\B{0.652}} & 0.424 & \U{\B{0.500}} & 0.524 & \U{\B{0.654}} \\
        & pr-AUC & 0.417 & \U{\B{0.543}} & 0.417 & \U{\B{0.688}} & 0.424 & \U{\B{0.699}} & 0.417 & \U{\B{0.815}} \\   \hline
        \multirow{3}{*}{$4$}  
        & $R^2_{test}$ & \B{0.854} & 0.850  & \B{0.888} & 0.886 & \B{0.458} & 0.451  & \B{0.484} & 0.481 \\
        & $R^2_{ind}$ & 0.440 & \U{\B{0.542}}  & 0.539 & \U{\B{0.686}} & 0.432 & U{\B{0.494}}  & 0.545 & \U{\B{0.652}} \\
        & pr-AUC & 0.513 & \U{\B{0.661}} & 0.514 & \U{\B{0.766}} & 0.517 & \U{\B{0.718}} & 0.514 & \U{\B{0.810}} \\   \hline
        \multirow{3}{*}{$5$}  
        & $R^2_{test}$ & 0.842 & \B{0.854}  & 0.902 & \B{0.904} & 0.437 & \B{0.441}  & 0.488 & \B{0.492} \\
        & $R^2_{ind}$ & 0.768 & \B{0.811}  & \B{0.956} & 0.951  & 0.692 & \B{0.719}  & \B{0.930} & 0.921 \\
        & pr-AUC & 0.728 & \U{\B{0.980}}  & 0.947 & \U{\B{1.000}} & 0.694 & \U{\B{0.943}}  & 0.946 & \U{\B{1.000}} \\   \hline
        \multirow{3}{*}{$6$}  
        & $R^2_{test}$ & {0.850} & \B{0.856}	 & 0.900 & \B{0.902} & 0.439 & \B{0.440} & 0.487 & \B{0.491} \\
        & $R^2_{ind}$ & 0.811 & \B{0.823} & \B{0.964} & 0.952 & \B{0.751} & 0.750 & \B{0.943} & 0.923 \\
        & pr-AUC & 1.000 & 1.000  & 1.000 & 1.000 & 1.000 & 1.000  & 1.000 & 1.000 \\   \hline
        \multirow{3}{*}{$7$}  
        & $R^2_{test}$ & 0.842 & \B{0.854}  & 0.901 & \B{0.904} & 0.427 & \B{0.430}  & 0.483 & \B{0.487} \\
        & $R^2_{ind}$ & 0.753 & \B{0.803} & 0.941 & \B{0.951} & 0.672 & \B{0.693} & \B{0.900} & 0.898 \\
        & pr-AUC & 0.753 & \U{\B{0.970}} & 0.927 & \U{\B{1.000}} & 0.744 & \U{\B{0.875}}  & 0.896 & \U{\B{1.000}} \\   \hline
    \end{tabular}
    \label{tab: 10, continuous}
\end{table}

\begin{table}[ht]
    \centering
    \caption{Same as Table \ref{tab: results discrete}, but for $P=50$ discrete features and for both configurations of the noise parameter $\phi$.}
    \begin{tabular}{|c|l||c|c|c|c|c|c|c|c|}\hline
        regression & evaluation &
        \multicolumn{4}{|c|}{$\phi = 0.1$} & \multicolumn{4}{|c|}{$\phi = 1.0$} \\ \cline{3-10}
        function & metric & 
        \multicolumn{2}{|c|}{$N = 500$} & \multicolumn{2}{|c|}{$N=5000$} &
        \multicolumn{2}{|c|}{$N = 500$} & \multicolumn{2}{|c|}{$N=5000$} 
        \\ \cline{3-10}
        &   & RF    & losaw & RF    & losaw & RF    & losaw & RF    & losaw \\ 
        &   &       & RF    &       & RF    &       & RF    &       & RF \\ \hline\hline
        \multirow{3}{*}{\hfil$1$}  
        & $R^2_{test}$ & 0.882 & \B{0.885} & 0.903 & \B{0.904} & \B{0.467} & 0.464  & \B{0.488} & 0.488 \\
        & $R^2_{ind}$ & 0.658 & \B{0.696}  & 0.943 & \B{0.951} & 0.687 & \B{0.697}  & \B{0.933} & 0.927\\
        & pr-AUC & 1.000 & 1.000 & 1.000 & 1.000 & 0.997 & \B{1.000} & 1.000 & 1.000\\   \hline
        \multirow{3}{*}{$2$}  
        & $R^2_{test}$ & 0.867 & \B{0.867}  & 0.904 & \B{0.906} & \B{0.463} & 0.452  & \B{0.481} & 0.480\\
        & $R^2_{ind}$ & 0.710 & \B{0.737}  & 0.905 & \B{0.923} & \B{0.661} & {0.624}  & \B{0.881} & 0.876\\
        & pr-AUC & \B{1.000} & 0.999 & 1.000  & 1.000 & \B{0.993} & 0.988 & 1.000  & 1.000\\   \hline
        \multirow{3}{*}{$3$}  
        & $R^2_{test}$ & 0.876 & \B{0.882} & 0.895 & \B{0.900} & \B{0.494} & 0.491 & 0.488 & \B{0.489}\\
        & $R^2_{ind}$ & 0.367 & \U{\B{0.552}} & 0.648 & \B{0.820} & 0.460 & \U{\B{0.628}} & 0.555 & \U{\B{0.748}}\\
        & pr-AUC & 0.417 & \U{\B{0.875}} & 0.417 & \U{\B{1.000}} & 0.545 & \U{\B{0.982}} & 0.417 & \U{\B{0.994}}\\   \hline
        \multirow{3}{*}{$4$}  
        & $R^2_{test}$ & 0.864 & \B{0.866}  & 0.898 & \B{0.901} & \B{0.451} & 0.449  & 0.498 & \B{0.499}\\
        & $R^2_{ind}$ & 0.419	 & \U{\B{0.475}}  & 0.636 & \U{\B{0.826}} & 0.455 & \U{\B{0.548}}  & 0.604 & \U{\B{0.734}}\\
        & pr-AUC & 0.515 & \B{0.524} & 0.514 & \U{\B{0.993}} & 0.602 & \U{\B{0.898}} & 0.514 & \U{\B{0.913}}\\   \hline
        \multirow{3}{*}{$5$}  
        & $R^2_{test}$ & 0.891 & \B{0.891}  & 0.906 & \B{0.907} & \B{0.467} & 0.459  & 0.498 & \B{0.498} \\
        & $R^2_{ind}$ & 0.773 & \B{0.814}  & 0.866 & \B{0.931} & 0.732 & \U{\B{0.787}}  & 0.871 & \U{\B{0.932}} \\
        & pr-AUC & 0.999 & \B{1.000}  & 1.000 & 1.000 & 0.855 & \U{\B{0.970}}  & 0.963 & \B{1.000}\\   \hline
        \multirow{3}{*}{$6$}  
        & $R^2_{test}$ & 0.882 & \B{0.886} & 0.905 & \B{0.906} & \B{0.469} & 0.465 & \B{0.497} & 0.496 \\
        & $R^2_{ind}$ & 0.778 & \U{\B{0.828}} & 0.947 & \B{0.965} & 0.750 & \B{0.771} & 0.945 & \B{0.948}\\
        & pr-AUC & 0.998 & \B{1.000}  & 1.000 & 1.000 & \B{0.995} & 0.994  & 1.000 & 1.000\\   \hline
        \multirow{3}{*}{$7$}  
        & $R^2_{test}$ & \B{0.879} & 0.878  & 0.903 & \B{0.904} & \B{0.466} & 0.457  & \B{0.478} & 0.477\\
        & $R^2_{ind}$ & 0.668 & \B{0.707} & 0.881 & \U{\B{0.940}} & \B{0.632} & 0.617 & 0.827 & \B{0.869}\\
        & pr-AUC & 0.737 & \U{\B{0.790}}  & 0.825 & \U{\B{1.000}} & 0.746 & \U{\B{0.818}}  & 0.791 & \U{\B{0.954}} \\   \hline
    \end{tabular}
    \label{tab: 50, discrete}
\end{table}

\begin{table}[ht]
    \centering
    \caption{Same as Table \ref{tab: results continuous}, but for $P=50$ continuous features and for both configurations of the noise parameter $\phi$.}
    \begin{tabular}{|c|l||c|c|c|c|c|c|c|c|}\hline
        regression & evaluation &
        \multicolumn{4}{|c|}{$\phi = 0.1$} & \multicolumn{4}{|c|}{$\phi = 1.0$} \\ \cline{3-10}
        function & metric & 
        \multicolumn{2}{|c|}{$N = 500$} & \multicolumn{2}{|c|}{$N=5000$} &
        \multicolumn{2}{|c|}{$N = 500$} & \multicolumn{2}{|c|}{$N=5000$} 
        \\ \cline{3-10}
        &   & RF    & losaw & RF    & losaw & RF    & losaw & RF    & losaw \\ 
        &   &       & RF    &       & RF    &       & RF    &       & RF \\ \hline\hline
        \multirow{3}{*}{\hfil$1$}  
        & $R^2_{test}$ & \B{0.871} & 0.869  & 0.898 & \B{0.898} & \B{0.451} & 0.444  & \B{0.487} & 0.487 \\
        & $R^2_{ind}$ & 0.680 & \B{0.716}  & 0.798 & \B{0.812} & 0.595 & \B{0.636}  & 0.766 & \B{0.797} \\
        & pr-AUC & 1.000 & 1.000 & 1.000 & 1.000 & \B{1.000} & 0.997 & 1.000 & 1.000 \\   \hline
        \multirow{3}{*}{$2$}  
        & $R^2_{test}$ & \B{0.834} & 0.829  & \B{0.887} & 0.885 & \B{0.434} & 0.424  & \B{0.480} & 0.476 \\
        & $R^2_{ind}$ & 0.676 & \B{0.705}  & 0.817 & \B{0.842} & 0.588 & \B{0.598}  & 0.778 & \B{0.795} \\
        & pr-AUC & 1.000 & 1.000 & 1.000  & 1.000 & \B{0.987} & 0.960 & 1.000  & 1.000 
 \\   \hline
        \multirow{3}{*}{$3$}  
        & $R^2_{test}$ & \B{0.854} & 0.851 & 0.888 & \B{0.890} & \B{0.450} & 0.439 & \B{0.483} & 0.482 \\
        & $R^2_{ind}$ & 0.343 & \U{\B{0.530}} & 0.462 & \U{\B{0.661}}  & 0.334 & \U{\B{0.484}} & 0.424 & \U{\B{0.646}} \\
        & pr-AUC & 0.417 & \U{\B{0.562}} & 0.417 & \U{\B{0.637}} & 0.424 & \U{\B{0.687}} & 0.417 & \U{\B{0.800}} \\   \hline
        \multirow{3}{*}{$4$}  
        & $R^2_{test}$ & \B{0.836} & 0.826  & \B{0.882} & 0.880 & \B{0.438} & 0.428  & \B{0.472} & 0.467 \\
        & $R^2_{ind}$ & 0.385 & \U{\B{0.519}}  & 0.478 & \U{\B{0.685}} & 0.371 & \U{\B{0.470}}  & 0.452 & \U{\B{0.633}} \\
        & pr-AUC & 0.513 & \U{\B{0.641}} & 0.514 & \U{\B{0.727}} & 0.511 & \U{\B{0.716}} & 0.514 & \U{\B{0.768}} \\   \hline
        \multirow{3}{*}{$5$}  
        & $R^2_{test}$ & 0.832 & \B{0.848}  & 0.901 & \B{0.903} & 0.431 & \B{0.441}  & 0.490 & \B{0.494}\\
        & $R^2_{ind}$ & 0.751 & \U{\B{0.813}}  & \B{0.968} & 0.965   & 0.660 & \U{\B{0.734}} & 0.938 & \B{0.940}   \\
        & pr-AUC & 0.642 & \U{\B{0.969}}  & 0.888 & \U{\B{1.000}} & 0.657 & \B{0.941}  & 0.882 & \U{\B{1.000}} \\   \hline
        \multirow{3}{*}{$6$}  
        & $R^2_{test}$ & 0.848 & \B{0.854} & 0.903 & \B{0.904} & \B{0.438} & 0.438 & 0.488 & \B{0.489} \\
        & $R^2_{ind}$ & 0.823 & \B{0.842} & \B{0.975} & 0.966 & 0.755 & \B{0.759} & \B{0.957} & 0.935\\
        & pr-AUC & 1.000 & 1.000  & 1.000 & 1.000 & \B{0.999} & 0.993  & 1.000 & 1.000 \\   \hline
        \multirow{3}{*}{$7$}  
        & $R^2_{test}$ & 0.833 & \B{0.847}  & 0.899 & \B{0.902} & 0.424 & \B{0.427}  & 0.483 & \B{0.487} \\
        & $R^2_{ind}$ & 0.746 & \U{\B{0.809}} & 0.947 & \B{0.961} & 0.642 & \B{0.685} & 0.902 & \B{0.912} \\
        & pr-AUC & 0.743 & \U{\B{0.971}}  & 0.920 & \U{\B{1.000}} & 0.742 & \U{\B{0.877}}  & 0.904 & \U{\B{1.000}} 
\\   \hline
    \end{tabular}
    \label{tab: 50, continuous}
\end{table}

\begin{table}[ht]
    \centering
    \caption{Same as Table \ref{tab: results discrete}, but for $P=100$ discrete features and for both configurations of the noise parameter $\phi$.}
    \begin{tabular}{|c|l||c|c|c|c|c|c|c|c|}\hline
        regression & evaluation &
        \multicolumn{4}{|c|}{$\phi = 0.1$} & \multicolumn{4}{|c|}{$\phi = 1.0$} \\ \cline{3-10}
        function & metric & 
        \multicolumn{2}{|c|}{$N = 500$} & \multicolumn{2}{|c|}{$N=5000$} &
        \multicolumn{2}{|c|}{$N = 500$} & \multicolumn{2}{|c|}{$N=5000$} 
        \\ \cline{3-10}
        &   & RF    & losaw & RF    & losaw & RF    & losaw & RF    & losaw \\ 
        &   &       & RF    &       & RF    &       & RF    &       & RF \\ \hline\hline
         \multirow{3}{*}{\hfil$1$}  
        & $R^2_{test}$ & 0.880 & \B{0.881}  & 0.904 & \B{0.904} & \B{0.481} & 0.476  & \B{0.489} & 0.487 \\
        & $R^2_{ind}$ & 0.705 & \B{0.713}  & 0.945 & 0.953 & 0.661 & \B{0.669}  & \B{0.939} & 0.920 \\
        & pr-AUC & 1.000 & 1.000 & 1.000 & 1.000 & 1.000 & 1.000 & 1.000 & 1.000 \\   \hline
        \multirow{3}{*}{$2$}  
        & $R^2_{test}$ & \B{0.874} & 0.867  & 0.897 & \B{0.898} & \B{0.474} & 0.461  & \B{0.482} & 0.481 \\
        & $R^2_{ind}$ & \B{0.718} & 0.703  & 0.907 & \B{0.923} & \U{\B{0.629}} & 0.551  & \B{0.870} & 0.868 \\
        & pr-AUC & 1.000 & 1.000 & 1.000  & 1.000 & \B{0.992} & 0.976 & 1.000  & 1.000 \\   \hline
        \multirow{3}{*}{$3$}  
        & $R^2_{test}$ & 0.885 & \B{0.886} & 0.898 & \B{0.902} & \B{0.450} & 0.446 & 0.491 & \B{0.493} \\
        & $R^2_{ind}$ & 0.265 & \U{\B{0.358}}  & 0.600 & \U{\B{0.787}} & 0.264 & \U{\B{0.405}} & 0.446 & \U{\B{0.658}} \\
        & pr-AUC & 0.417 & 0.417 & 0.417 & \U{\B{0.999}} & 0.417 & \U{\B{0.593}} & 0.417 & \U{\B{0.959}} \\   \hline
        \multirow{3}{*}{$4$}  
        & $R^2_{test}$ & \B{0.864} & 0.863  & 0.897 & \B{0.901} & \B{0.483} & 0.477  & 0.490 & \B{0.491} \\
        & $R^2_{ind}$ & 0.306 & \U{\B{0.378}}  & 0.592 & \U{\B{0.799}} & 0.342 & \U{\B{0.406}}  & 0.576 & \U{\B{0.730}} \\
        & pr-AUC & 0.485 & \B{0.510} & 0.514 & \U{\B{0.961}} & 0.510 & \U{\B{0.631}} & 0.514 & \U{\B{0.905}} \\   \hline
        \multirow{3}{*}{$5$}  
        & $R^2_{test}$ & 0.868 & \B{0.871}  & 0.906 & \B{0.908} & \B{0.470} & 0.459  & \B{0.496} & 0.496\\
        & $R^2_{ind}$ & 0.656 & \B{0.693}  & 0.899 & \U{\B{0.959}} &  0.761 & \B{0.783}  & 0.864	 & \B{0.912}   \\
        & pr-AUC & 0.571 & \U{\B{0.797}}  & 0.702 & \U{\B{1.000}} &0.938 & \B{0.964}  & 0.951 & \B{1.000} \\   \hline
        \multirow{3}{*}{$6$}  
        & $R^2_{test}$ & 0.873 & \B{0.875} & 0.906 & \B{0.907} & \B{0.473} & 0.466 & \B{0.494} & 0.493 \\
        & $R^2_{ind}$ & 0.780 & \B{0.810} & 0.961 & \B{0.972} & \B{0.752} & 0.730 & 0.952 & \B{0.954}\\
        & pr-AUC & 1.000 & 1.000  & 1.000 & 1.000 & \B{1.000} & 0.998  & 1.000 & 1.000  \\   \hline
        \multirow{3}{*}{$7$}  
        & $R^2_{test}$ & \B{0.879} & 0.879  & 0.905 & \B{0.906} & \B{0.455} & 0.446  & \B{0.502} & 0.502 \\
        & $R^2_{ind}$ & 0.695 & \B{0.738} & 0.871 & \U{\B{0.932}} & 0.644 & \B{0.646} & 0.849 & \B{0.884} \\
        & pr-AUC & 0.696 & \U{\B{0.797}}  & 0.840 & \U{\B{1.000}} & 0.868 & \U{\B{0.947}}  & 0.826 & \U{\B{0.982}} \\   \hline
    \end{tabular}
    \label{tab: 100, discrete}
\end{table}

\begin{table}[ht]
    \centering
    \caption{Same as Table \ref{tab: results continuous}, but for $P=100$ continuous features and for both configurations of the noise parameter $\phi$.}
    \begin{tabular}{|c|l||c|c|c|c|c|c|c|c|}\hline
        regression & evaluation &
        \multicolumn{4}{|c|}{$\phi = 0.1$} & \multicolumn{4}{|c|}{$\phi = 1.0$} \\ \cline{3-10}
        function & metric & 
        \multicolumn{2}{|c|}{$N = 500$} & \multicolumn{2}{|c|}{$N=5000$} &
        \multicolumn{2}{|c|}{$N = 500$} & \multicolumn{2}{|c|}{$N=5000$} 
        \\ \cline{3-10}
        &   & RF    & losaw & RF    & losaw & RF    & losaw & RF    & losaw \\ 
        &   &       & RF    &       & RF    &       & RF    &       & RF \\ \hline\hline
        \multirow{3}{*}{\hfil$1$}  
        & $R^2_{test}$  & \B{0.871} & 0.869  & 0.901 & \B{0.901}   & \B{0.454} & 0.446  & \B{0.487} & 0.487 \\
        & $R^2_{ind}$   & 0.686& \B{0.723}  & 0.806 & \B{0.822} & 0.591 & \B{0.635}  & 0.768 & \B{0.802}\\
        & pr-AUC        & 1.000  & 1.000 & 1.000 & 1.000  & \B{1.000} & 0.997 & 1.000 & 1.000 \\   \hline
        \multirow{3}{*}{$2$}  
        & $R^2_{test}$  & \B{0.830} & 0.823  & \B{0.889} & 0.887 & \B{0.425} & 0.416  & \B{0.481} & 0.477 \\
        & $R^2_{ind}$   & 0.681	 & \B{0.709}  & 0.823 & \B{0.849} & 0.582 & \B{0.587}  & 0.771 & \B{0.797} \\
        & pr-AUC        & \B{1.000} & 0.999  & 1.000 & 1.000 & \B{0.988} & 0.951 & 1.000  & 1.000 \\   \hline
        \multirow{3}{*}{$3$}  
        & $R^2_{test}$  & \B{0.849} & 0.845  & 0.885 & \B{0.888} & \B{0.447} & 0.436 & \B{0.479} & 0.478\\
        & $R^2_{ind}$   & 0.315 & \U{\B{0.516}}  & 0.445 & \U{\B{0.668}} & 0.313 & \U{\B{0.476}} & 0.381 & \U{\B{0.648}}\\
        & pr-AUC        & 0.417 & \U{\B{0.547}} & 0.417 & \U{\B{0.656}} & 0.421 & \U{\B{0.678}} & 0.417 & \U{\B{0.786}} \\   \hline
        \multirow{3}{*}{$4$}  
        & $R^2_{test}$  & \B{0.833} & 0.823  & \B{0.879} & 0.877 & \B{0.440} & 0.430  & \B{0.478} & 0.475 \\
        & $R^2_{ind}$   & 0.369 & \U{\B{0.520}}  & 0.464 & \U{\B{0.685}} & {0.345} & \U{\B{0.463}}  & {0.423} & \B{0.635} \\
        & pr-AUC        & 0.513 & \U{\B{0.659}}  & 0.514 & \U{\B{0.714}} & 0.511 & \U{\B{0.716}} & 0.514 & \U{\B{0.787}}\\ \hline
        \multirow{3}{*}{$5$}  
        & $R^2_{test}$  & 0.830 & \B{0.848}  & 0.902 & \B{0.903} & 0.431 & \B{0.441}  & 0.488 & \B{0.490} \\
        & $R^2_{ind}$   & 0.752 & \U{\B{0.815}}  & \B{0.972} & 0.969 & 0.661 & \U{\B{0.743}} & {0.941} & \B{0.945} \\
        & pr-AUC        & 0.629 & \U{\B{0.959}} & 0.881 & \U{\B{1.000}}  & 0.677 & \U{\B{0.955}}  & 0.856& \U{\B{1.000}} \\   \hline
        \multirow{3}{*}{$6$}  
        & $R^2_{test}$  & {0.844} & \B{0.850}  & 0.902 & \B{0.903} & {0.440} & \B{0.441} & 0.495 & \B{0.497}\\
        & $R^2_{ind}$   & {0.826}	 & \B{0.846}  & \B{0.979} & 0.970  & {0.757} & \B{0.775} & \B{0.961} & 0.942\\
        & pr-AUC        & 1.000 & 1.000 & 1.000 & 1.000 &  \B{0.999} & 0.996  & 1.000 & 1.000 \\   \hline
        \multirow{3}{*}{$7$}  
        & $R^2_{test}$  & 0.829 & \B{0.843}  & 0.898 & \B{0.901} & 0.417 & \B{0.421}  & 0.483 & \B{0.487}\\
        & $R^2_{ind}$   & 0.741 & \U{\B{0.805}}  & 0.949 & \B{0.964} & 0.623 & \U{\B{0.682}} & 0.901 & \B{0.915} \\
        & pr-AUC        & 0.734 & \U{\B{0.958}}  & 0.919 & \U{\B{1.000}} & 0.736 & \U{\B{0.885}}  & 0.896 & \U{\B{1.000}} \\   \hline
    \end{tabular}
    \label{tab: 100, continuous}
\end{table}
\end{document}